\documentclass[10pt,twocolumn,letterpaper]{article}

\usepackage[accsupp]{axessibility}  

\usepackage[pagenumbers]{wacv} 

\usepackage{graphicx}
\usepackage{amsmath}
\usepackage{amssymb}
\usepackage{booktabs}
\usepackage{subfiles}

\usepackage{pifont}
\newcommand{\cmark}{\ding{51}}%
\newcommand{\xmark}{\ding{55}}%

\newcommand{\cad}{{CompAD}}

\usepackage[pagebackref,breaklinks,colorlinks]{hyperref}

\usepackage[capitalize]{cleveref}
\crefname{section}{Sec.}{Secs.}
\Crefname{section}{Section}{Sections}
\Crefname{table}{Table}{Tables}
\crefname{table}{Tab.}{Tabs.}

\def\wacvPaperID{1327} 
\def\confName{WACV}
\def\confYear{2024}

\begin{document}

\title{A Hybrid Graph Network for Complex Activity Detection in Video}


\author{%
Salman Khan$^1$ \quad Izzeddin Teeti$^1$ \quad Andrew~Bradley$^1$ \quad Mohamed Elhoseiny$^2$ Fabio~Cuzzolin$^1$\\
$^1$Oxford Brookes University \quad $^2$KAUST \\
{\tt\small\{salmankhan,  fabio.cuzzolin\}@brookes.ac.uk}
}

\maketitle

\begin{abstract}
\vspace{-2mm}
Interpretation and understanding of video presents a challenging computer vision task in numerous fields - e.g. autonomous driving and sports analytics. Existing approaches to interpreting the actions taking place within a video clip are based upon Temporal Action Localisation (TAL), which typically identifies short-term actions. The emerging field of \textbf{Comp}lex \textbf{A}ctivity \textbf{D}etection (\cad{}) extends this analysis to long-term activities, with a deeper understanding obtained by modelling the internal structure of a complex activity taking place within the video.



We address the \cad{} problem using a hybrid graph neural network which combines attention applied to a graph encoding the local (short-term) dynamic scene with a temporal graph modelling the overall long-duration activity. Our approach is as follows: i) Firstly, we propose a novel feature extraction technique which, for each video snippet, generates spatiotemporal `tubes' for the active elements (`agents') in the (local) scene by detecting individual objects, tracking them and then extracting 3D features from all the agent tubes as well as the overall scene. ii) Next, we construct a local scene graph where each node (representing either an agent tube or the scene) is connected to all other nodes. Attention is then applied to this graph to obtain an overall representation of the local dynamic scene. iii) Finally, all local scene graph representations are interconnected via a temporal graph, to estimate the complex activity class together with its start and end time.

The proposed framework outperforms all previous state-of-the-art methods on all three datasets including ActivityNet-1.3, Thumos-14, and ROAD.
\end{abstract}

\vspace{-4mm}
\section{Introduction} \label{sec:introduction}
\vspace{-1mm}


{Detecting and recognising activities in untrimmed videos is a challenging research problem, with applications to, e.g.,}
sports \cite{hu2020progressive}, autonomous driving \cite{caesar2020nuscenes}, medical robotics \cite{xia2023nested} and surveillance \cite{yuan2017temporal}. 
{
\emph{Temporal Action Localisation} (TAL) approaches not only recognise the action label(s) present in a video, but can also 
identify the start and end time of each activity instance, enabling the
generation} of sports highlights \cite{li2021multisports,zeng2021graph}, the understanding of road scenes in autonomous driving \cite{skhan2021comp}, the video summarisation 
of surveillance videos \cite{xiao2020convolutional} and video captioning \cite{krishna2017dense,mun2019streamlined}.
A number of TAL methods \cite{lin2018bsn,zeng2019graph,lin2019bmn,long2019gaussian,liu2019multi,xu2020g,xia2022learning,liu2022empirical,hsieh2022contextual,bao2022opental} have recently been proposed, competing to achieve state-of-the-art performance \cite{cheng2022tallformer,zhang2022actionformer} on accepted benchmarks.
Whereas various new datasets have been recently proposed, 
the two most common relevant benchmarks 
remain ActivityNet 1.3 \cite{caba2015activitynet}, and Thumos-14 \cite{idrees2017thumos}. 
State-of-the-art performance on Thumos-14 has improved in four years by some 19\% \cite{zhu2022learning}.

\begin{figure*}[h]
    \centering
    \includegraphics[width=0.85\textwidth]{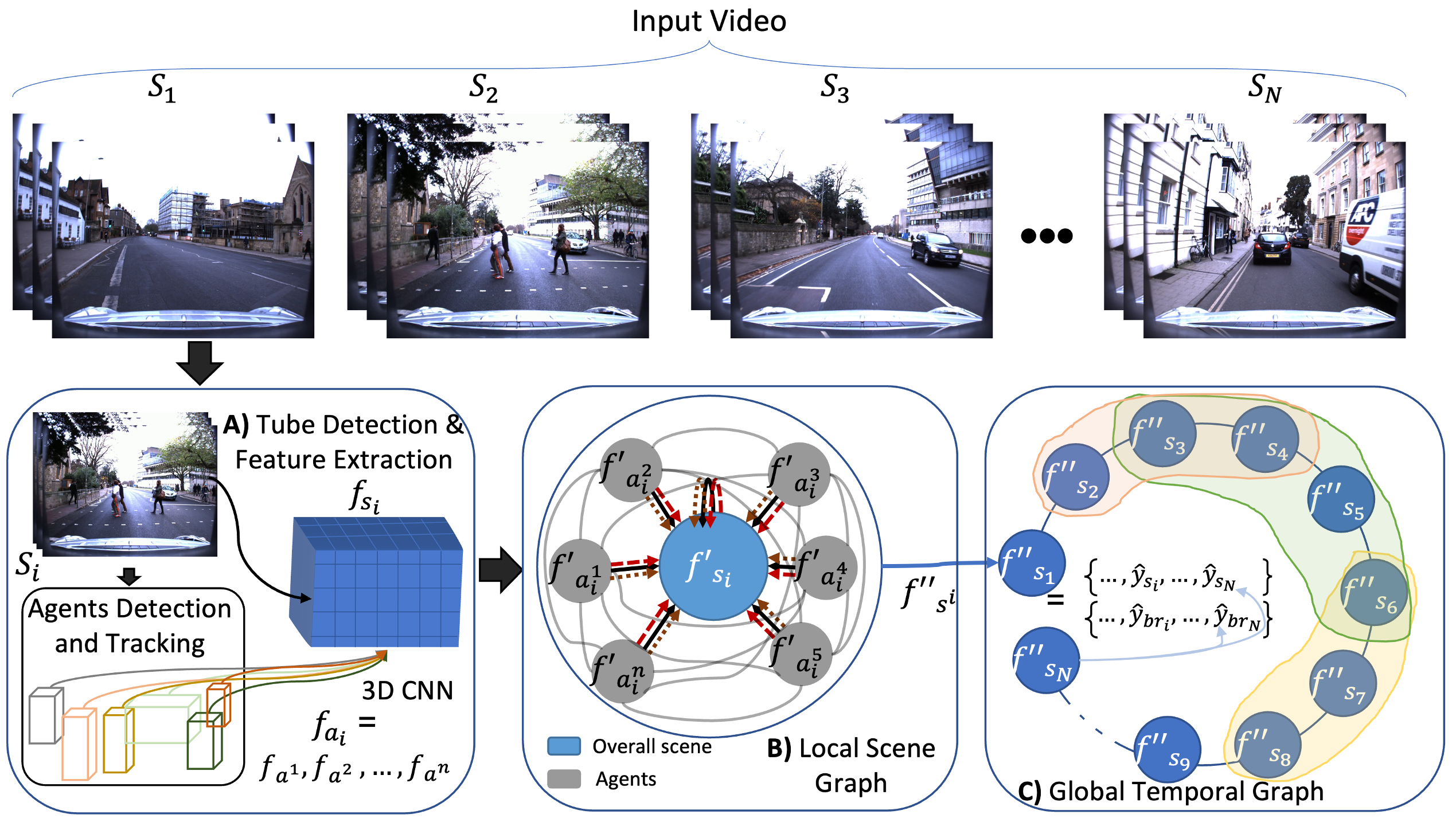}
    \vspace{-2mm}
    \caption{An overview of our \textbf{Comp}lex \textbf{A}ctivity \textbf{D}etection (\cad{}) framework. The input video is divided into fixed-size snippets $S_1,\ldots,S_N$ (top); each snippet is then processed in three major steps (bottom). A) Firstly, scene objects (agents) are detected and tracked throughout the snippet to form agent tubes. 3D features are then extracted from all the cropped agent tubes ($f_{a^i}$) as well as the local scene ($f_{s_i}$). B) Next, a 
    local scene graph is constructed where agent nodes (in gray) are connected to each other and to the snippet node (in light blue). The local scene graph is processed using a graph attention
    network (GAT), resulting in intermediate scene features ($f''_{s_i}$). 
    C) {Finally, 
    all local scene features associated with individual snippets are temporally connected 
    and processed using a global temporal graph to identify the boundaries of the activity 
    using anchor proposals (shown in different colors). 
    }
    }
    
    \label{fig:framework}
    \vspace{-5mm}
\end{figure*}

Almost all TAL approaches 
comprise a
\emph{features/scene representation} stage and a \emph{temporal localisation} stage.
In the 
former, snippets (continuous sequences of frames) are processed to understand the local scene in the video.
Methods \cite{xia2022learning,liu2022empirical,hsieh2022contextual,bao2022opental} typically employ pre-extracted features obtained using a  sequential learning model (e.g., I3D \cite{carreira2017quo}), often pre-trained on the Kinetics \cite{kay2017kinetics} dataset. 
Features are then processed, 
e.g., 
via a temporal or semantic graph neural network, by applying appropriate encoding techniques 
or by generating temporal proposals, in an object detection style \cite{chao2018rethinking}.
In the second stage, 
{TAL approaches temporally localise activities in various ways,}
e.g. via temporal graphs~\cite{xu2020g,zeng2019graph}, boundary regression and proposals generation \cite{lin2018bsn,lin2019bmn,hsieh2022contextual} or encoder-decoder methods \cite{zhu2022learning}. 

{
As recently pointed out in e.g. \cite{khan2021spatiotemporal,cheng2022tallformer}, in real-world applications a challenge is posed by \emph{complex activities}, longer-term events comprising a series of elementary actions, often performed by multiple agents.}
For example, an Autonomous Vehicle (AV) negotiating a pedestrian crossing is engaged in a complex activity: First it drives along the road, then the traffic lights change to red, the vehicle stops and several pedestrians cross the road. Eventually, the lights turn green again and the AV drives off.

\begin{figure}[h]
    \centering
    \vspace{-2mm}
    \includegraphics[width=0.45\textwidth]{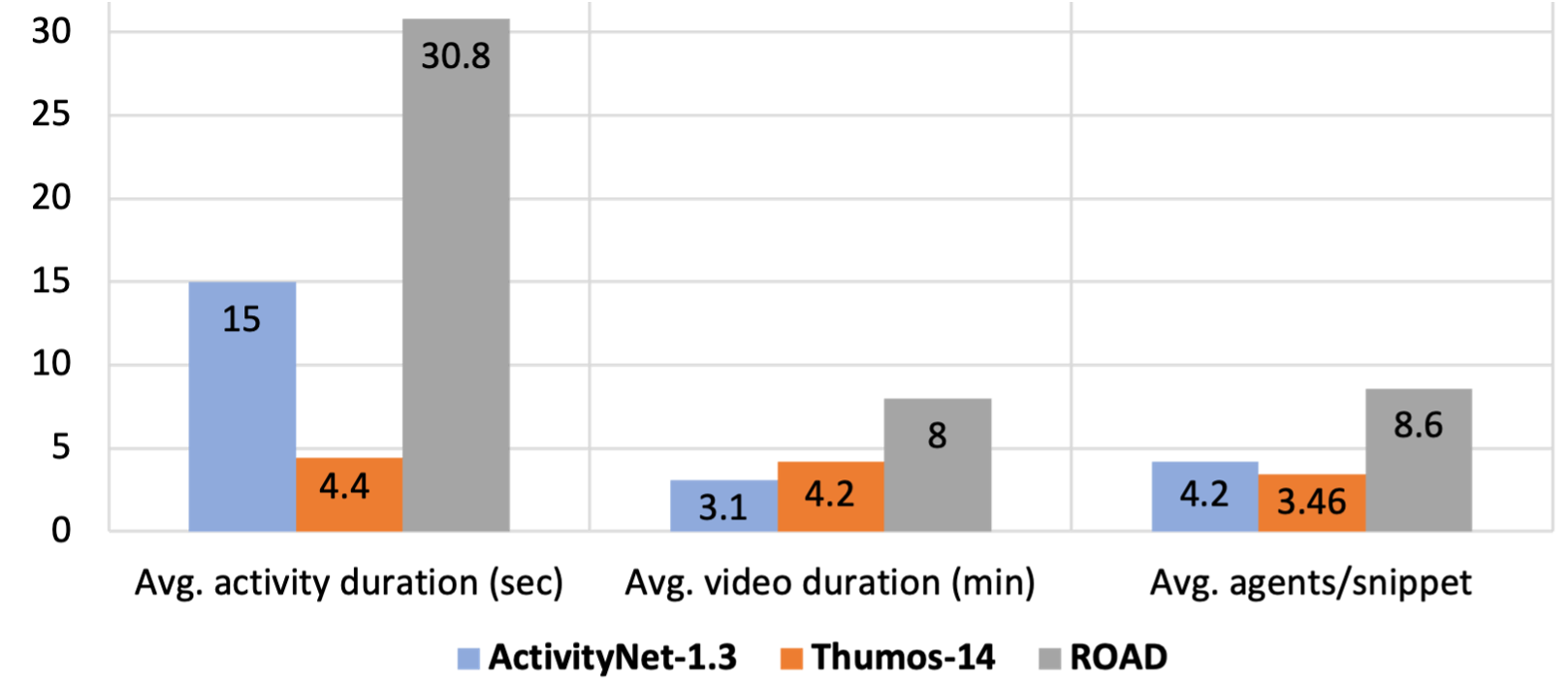}
    \vspace{-2mm}
    \caption{{ Comparing TAL datasets (ActivityNet-1.3 and Thumos-14) with a CompAD dataset (ROAD) in terms of average activity and video durations and mean number of agents per snippet.}
    }
    \label{fig:comp}
    \vspace{-2mm}
\end{figure}

{
Theoretically, TAL methods can be employed to temporally segment complex activities, in practice such approaches are only employed to detect short- or mid-duration actions lasting a few seconds at most ({e.g., a person jumping or pitching a baseball}). The activities contemplated by the most common datasets are of this nature.
Fig. \ref{fig:comp} compares two standard TAL benchmarks with the recently released ROAD dataset, explicitly designed for complex activity detection. 
Activities in the ROAD dataset last longer than those in ActivityNet or Thumos, with twice as many agents per snippet, making them more complex in nature.}

{
In this paper we argue that standard TAL approaches are ill equipped to detect complex activities, as they fail to model both the global temporal structure of the activity and its fine-grained structure, in terms of the agents and elementary actions involved and their relations.
}

{
We may thus define (strong) \emph{\textbf{Comp}lex \textbf{A}ctivity \textbf{D}etection} (\cad{}) as the task of recovering, given an input video, the temporal extent of the activities there present, \emph{as well as} their inner structure in terms of the agents or elementary actions involved. A weaker \cad{} is one in which the only expected output is the temporal segmentation,
with the internal structure of the activity estimated as a means for achieving segmentation - with no annotation available.

}

While a small number of studies have attempted to detect complex, long-duration activities \cite{khan2021spatiotemporal,cheng2022tallformer,shou2016temporal}, 
to our knowledge \cite{skhan2021comp} is the only existing study which attempts to tackle \cad{} as defined above.
The work, however, relies upon the availability of 
heavily-annotated datasets which include granular labels for the individual actions which make up a complex activity, and the corresponding frame-level bounding boxes.
This is a serious limitation, for it prevents  \cite{skhan2021comp} from being usable for pure temporal segmentation and compared with prior art there. 
{Further, \cite{skhan2021comp} focuses only on the graph representation of snippets, neglecting the long-term modelling of complex activities.}

 \vspace{-1mm}
\textbf{Objectives}. 
This work aims to push the boundaries of temporal action localisation to tackle (weak) \cad{},
leveraging datasets providing temporal segmentation annotation \emph{only}.
We do so by modelling and leveraging a complex video activity's internal structure, but
without resorting to any additional fine-grained annotation concerning individual actions. 
{Nevertheless, our proposal is easily generalisable to strong \cad{} whenever individual action/agent annotation is available, in an end-to-end trainable setting.}

Our \textbf{proposed framework} (Fig. \ref{fig:framework}) is composed of three stages: A) 
feature extraction;
B) a scene graph attention network designed to learn the importance of each active object (`agent') within the local dynamic scene; and C) a temporal graph of attended 
scene graphs for the localisation of complex activities of arbitrary duration.


Our {feature extraction} scheme (A) differs from what typically done in 
prior art
- where spatiotemporal features are extracted from whole snippets only. In contrast, we first detect the relevant active objects {(\emph{agents})} in the scene and track them within each snippet to build for each an \emph{agent tube} (a series of related bounding boxes).
We use a pre-trained tracker 
to allow the method to be deployable to any datasets with only temporal annotation (no bounding box annotation for the scene agents is required), while leveraging a fine-grained description of the internal structure of a complex activity {in the form of a graph of agents.}
A pre-trained 3D feature extraction model is used to extract features from both the agent tubes and the whole snippet.

To represent the local dynamic scene within each snippet, 
a \emph{local scene graph} (B) is constructed using three different topologies 
{(Sec. \ref{sec:scene-graph-attention}).}
The local scene graph is then processed using a \emph{scene graph attention (SGAT) network} to extract an overall scene representation, 
because of
its ability to compute the importance of each node in the context of its neighbors, thus modelling the structure of complex scenes.

Finally (C), the learned local scene graphs are connected to each other by constructing a \emph{temporal graph}, with the aim of recognising the activity label and identifying its temporal boundaries (start and end time) {in a class-agnostic manner}.

Our main \textbf{Contributions} are, therefore:
\begin{itemize}
\vspace{-2mm}
\item 
An original 
\emph{hybrid graph network} approach for general complex activity detection, comprising a local scene graph as well as a global temporal graph, capable of localising both complex and shorter-term activities 
{
and able to perform both weak and strong \cad{}, depending on what annotation is available.}
\begin{itemize}
\vspace{-2mm}
\item
A scene graph attention network for learning the importance of each agent in the context of a (local) dynamic scene. 
\vspace{-1mm}
\item 
A temporal graph of activated scene graphs for the detection of the start and end of an activity of arbitrary duration.
\end{itemize}
\item
\vspace{-2mm}
{
Comprehensive experiments showing how our approach leveraging weak \cad{} \emph{outperforms the most recent TAL state-of-the-art across the board} on ActivityNet-1.3, Thumos-14 and the recent ROad event Awareness Dataset (ROAD) \cite{singh2022road,skhan2021comp}, which portrays long-duration road activities involving multiple road agents over sometimes several minutes, showing clear dominance on classical TAL approaches.}
\end{itemize}

\vspace{-4mm}
\section{Related Work}
\vspace{-1mm}

{Recently, a series of works on activity detection have been proposed including spatiotemporal methods and graph-based methods, with recent advances in Graph Attention showing promise in many applications, including trajectory prediction for autonomous driving \cite{kosaraju2019social, izzeddin2022}, social recommendation \cite{song2019session} and 3D object tracking \cite{Buchner20223DMT}. The state-of-the-art approaches to activity detection and graph approaches are summarised below.}
\vspace{-1mm}
\subsection{One-stage vs two-stage approaches}
\vspace{-1mm}
TAL methods can been broadly divided into \emph{one-stage} and \emph{two-stage} approaches. 
The former
\cite{buch2019end,huang2019decoupling,long2019gaussian,lin2021learning} detect actions/activities in a single shot,
and 
can be easily trained in an end-to-end manner. For example, Wang et al. \cite{wang2021oadtr} detect actions using transformers which, unlike RNNs, do not suffer from nonparallelism and vanishing gradients. 
Most one-stage methods, however, only perform action classification, rather than spatiotemporal localisation. In contrast, Lin et al. \cite{lin2021learning} have recently proposed an anchor-free, one-stage light model which generates proposals, locates actions within them, and classifies them end-to-end.

The latter group of methods
\cite{lin2018bsn,lin2019bmn,bai2020boundary, Bao_2022_CVPR,Zhu_Wang_Tang_Liu_Zheng_Hua_2022}, instead, consist of two stages, similarly to region-proposal object detectors. The first stage generates suitable proposals for predicting the start and end time of an activity,
while the second stage extracts features and processes the proposals before passing them to both a classification head and a regression head (for temporal localisation). Some works, including \cite{lin2019bmn}, focus on the first stage to improve the quality of the proposals, while others focus on the processing or refining of the proposals. \cite{Zhu_Wang_Tang_Liu_Zheng_Hua_2022}, for instance, uses an off-the-shelf method for proposal generation. The second stage consists of two networks, a `disentanglement' network to separate the classification and regression representations, and a `context aggregation' network to add them together. Such methods are not trainable end-to-end and limited to short-or mid-duration actions.

In contrast, this paper proposes
a \cad{} method 
which exhibits the advantages
of both classes of methods, 
thanks to our hybrid graph approach capable of
recognising and localising both short- and long-duration activities.
\vspace{-1mm}
\subsection{Graph Convolutional Networks approaches}
\vspace{-1mm}
Graph Convolutional Networks (GCNs) have been extensively investigated for TAL \cite{zeng2019graph,bai2020boundary,xu2020g,nawhal2021activity,yang2022acgnet}. 
GCN-based TAL methods can also be further divided into two-stage and one-stage methods. 
The former, once again, perform localisation after generating suitable proposals.
E.g., in 
\cite{zeng2019graph} 
two different types of boundary proposals are generated and then individually passed to the same graph, 
resulting in both an action label and a temporal boundary. 

One-stage GCN-based TAL methods, instead, solve the detection problem without proposals in one go by learning 
spatiotemporal features in an end-to-end manner. For examples, 
in \cite{xu2020g} 
a graph is first generated by connecting the snippets both temporally 
and by virtue of meaningful semantics.
{This graph is then divided into sub-graphs (anchors), where each anchor represents the activity in an untrimmed video.}
{In contrast, \cite{skhan2021comp} proposed a spatiotemporal scene graph-based long-term TAL method where each of the snippets is considered as a separate graph, which is heavily dependent on the particular actions present in the scene, and is only applicable to datasets providing (label and bounding box) annotations for each individual action.}

This study proposes leveraging GCNs, by incorporating them in an overall hybrid graph capable of modelling both the local scenes, via a Graph Attention Network (GAT) \cite{velivckovic2017graph}, and 
the overall global activity via a temporal graph.
GATs build on the transformer concept by applying attention to graphs, and were originally proposed in \cite{velivckovic2017graph} 
for node classification. 
The idea is to update the representation of the current node with respect to its neighbours by applying attention to learn the importance of the various connections.

In this paper, 
GAT is used at the local scene level to learn the features of each node (active agent tube), to generate a more robust local scene representation. 


\vspace{-1mm}
\section{Proposed Methodology} \label{sec:methodology}
\vspace{-1mm}

The proposed framework
is illustrated in Fig. \ref{fig:framework}. An input untrimmed video $V$ is divided into $N$ snippets $S$ = $S_1$, ..., $S_i$, $S_{i+1}$, ..., $S_N$ (each snippet is a pre-defined constant length of consecutive frames). 
Each snippet is then passed to the tube detection and feature extraction module which returns a feature vectors for both the snippet $f_{s_i}$ and the individual agent tubes $f_{a^1_i}$, $f_{a^2_i}$,...,$f_{a^n_i}$, where $n$ is the number of agents present in snippet $i$. 
These features are then forwarded to the local scene graph attention layer for learning the attention of each agent in the context of its neighbours. This returns an aggregate feature representation for the whole scene ($f''_{s_i}$). These aggregate local scene features for all the snippets, $f''_{s_1}$,$f''_{s_2}$,...,$f''_{s_N}$, are then connected
using a global temporal graph for the generation of the activity class label $\hat{y}_{s_i}$ and {activity boundary labels $\hat{y}_{br_i}$ using anchor proposals in a class-agnostic manner.} 

\begin{figure}[h]
    \centering
    \includegraphics[width=0.40\textwidth]{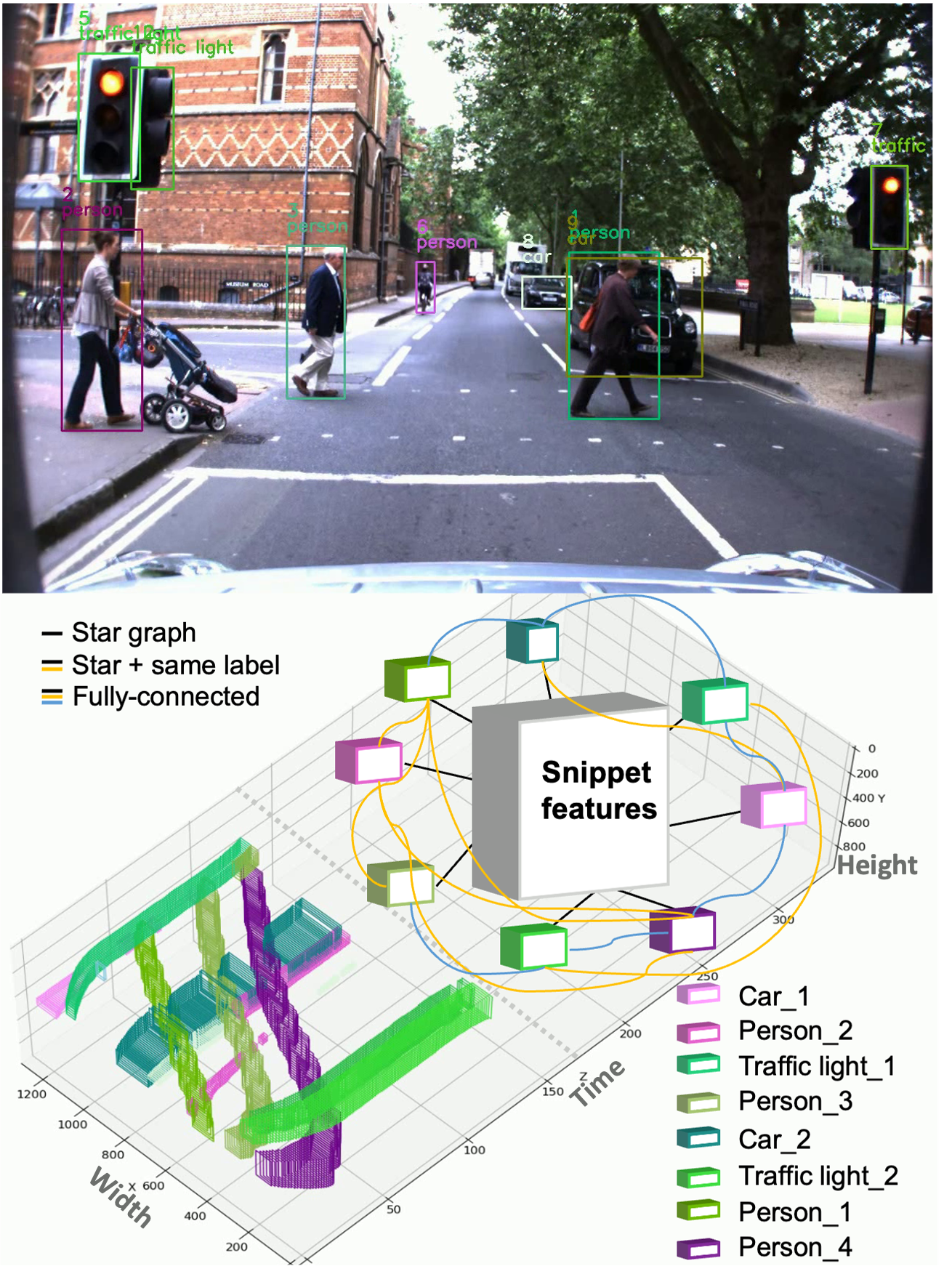}
    \vspace{-2mm}
    \caption{{ Visualisation of our agent detection and tracking stage using a bird's-eye view of the spatiotemporal volume corresponding to a video segment of the ROAD dataset. The upper section shows a random frame from the segment (with bounding boxes). Below, the detected agent tubes are plotted in space and time together with different possible local scene graph representations. The agent tubes we extract from the scene are of variable sizes. The way scene object motion affects the spatial and temporal extent of the tubes can be appreciated. Additional scenarios with visualisation are illustrated in the \textbf{Supplementary material.}}}
    \label{fig:tracking}
    \vspace{-5mm}
\end{figure}

\begin{figure*}[h]

    \centering
    \includegraphics[width=0.78\textwidth]{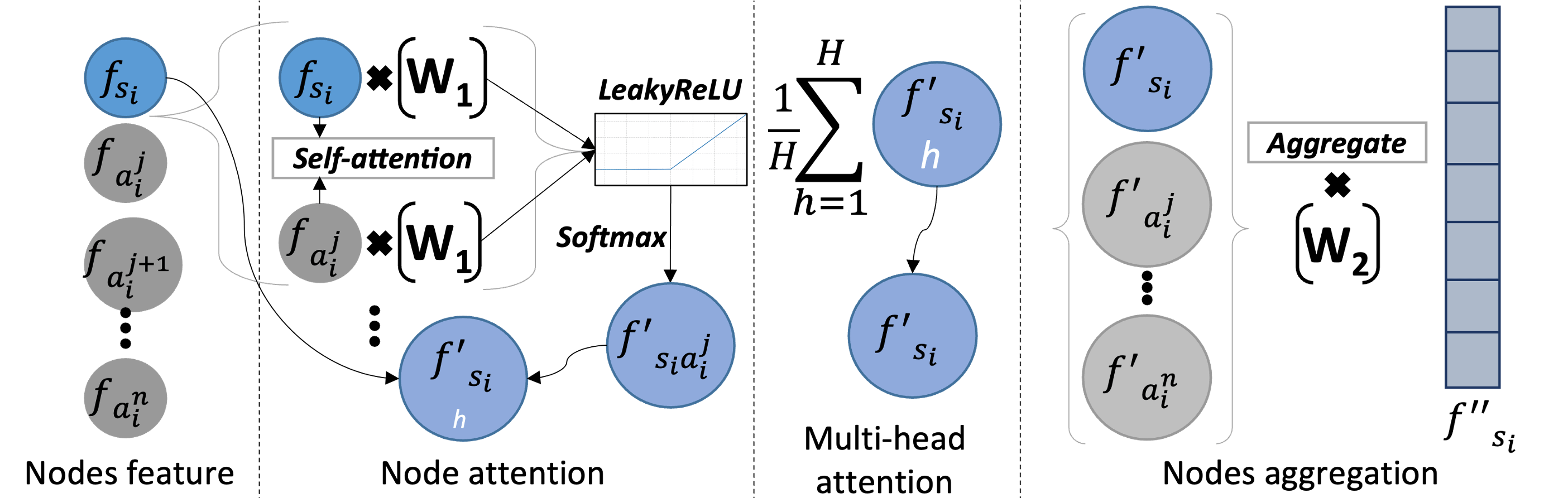}
    \vspace{-3mm}
    \caption{Our scene graph attention layer (Stage B of our approach, Fig. \ref{fig:framework}) takes the node features generated in the feature extraction Stage A as an input, and updates each node's features using node attention with respect to its all connected neighbours. Multi-head attention with $H$ heads is applied to each node to further robustify the representation -- averaging yields the final node features. To obtain a fixed-size overall representation for a specific local scene (snippet), the features of all its nodes are aggregated using a learnable weight matrix $W_2$. 
    }
    \label{fig:sgat}
    \vspace{-6mm}
\end{figure*}

\vspace{-1mm}
\subsection{Tube Detection and Feature Extraction} \label{sec:feature-extraction}
\vspace{-1mm}
As mentioned, one of the contributions of this paper is a new strategy for feature extraction and representation which consists in analysing the finer structure of the local dynamic scene, rather extracting features from whole snippets only.  

\textbf{Objects Detection and Tracking}. We first detect scene objects in each frame of the snippet using an object detector pre-trained on the COCO dataset \cite{lin2014microsoft}, which comprises of 80 different types of objects. However, we select a relevant list of object types (\emph{agents}) which depends on the dataset (the lists of classes selected for each dataset is given in Section \ref{sec:implem} -- Implementation Details). The detected agents are then tracked using a pre-trained tracker throughout the snippet in order to construct agent {tubes}. 
The latter are of variable length, depending on the agent itself and its role in each snippet. Fig. \ref{fig:tracking} pictorially illustrates agent tubes in an example video segment from the ROAD dataset, showing both a sample frame, with overlaid the detection bounding boxes, and a bird's-eye view of the agent tubes and local scene graph for the snippet it belongs to. 

\textbf{Feature Extraction}. Next, all the detected agent tubes are 
brought to a standard size
and passed to the pre-trained 3D CNN model along with the whole snippet for spatiotemporal feature extraction. The adopted 3D CNN model allows variable length inputs, and outputs a fixed-sized feature vector for each of the agent tubes and the snippet.


\vspace{-1mm}
\subsection{Scene Graph Attention} \label{sec:scene-graph-attention}
\vspace{-1mm}
In the second stage, a scene graph representation is used to describe the scene in terms of features extracted from both the overall snippet and the agent tubes.

\textbf{Graph Generation}. {The scene graph is constructed using three different topologies: a star graph connecting each agent node to the scene node, a star topology with also links between agents sharing a label, and a fully-connected one (Fig. \ref{fig:tracking}). 
The 
influence of topology is shown
in Sec. \ref{sec:ablation}.}


\textbf{Graph Attention}. As mentioned, our scene graph attention layer is inspired by the GAT concept \cite{velivckovic2017graph}, originally designed for node classification. While we follow a similar attention mechanism, here we amend the attention layer in order to extract aggregate features from all the nodes, to be passed in turn to our localisation layer in the third stage. 

The workflow of our scene graph attention layer is shown in Fig. \ref{fig:sgat}. All input node features are linearly transformed using a {weight matrix} $W_1$, followed by \emph{self-attention} to find the importance of each node with respect to its connected neighbours. An activation function LeakyReLU is applied to the features for nonlinearity, resulting in final output features for all of the nodes. These are then normalised using a softmax function to make each node representation ($f'_{{s_i}{a}^j_i}$) comparable with that of all the nodes connected to it ($f'_{s_i}$).

The {self-attention} process is further improved by applying a \emph{multi-head attention} strategy inspired by 
transformers \cite{velivckovic2017graph}.
Attention is applied to the node features individually. For each node, the average over the $H$ heads is computed, resulting in an attended feature vector for each node.  

Finally, to get a fixed-size representation for the whole scene, the output features of all the nodes are aggregated using another learnable weight matrix $W_2$, which outputs the final feature representation $f''_{s_i}$ for the whole (local) scene. 
\vspace{-1mm}
\subsection{Temporal Graph Localisation}
\label{sec:temporal-graph}
\vspace{-1mm}
In Stage C of our framework, activity recognition and localisation are performed using a GCN. The final features from all the local scenes  ($f''_{s_1}$, ..., $f''_{s_i}$,$f''_{s_{i+1}}$,...,$f''_{s_N}$), as outputted by the scene graph attention layer, are temporally connected to build a global temporal graph (see Fig. \ref{fig:framework}, C).

Our GCN network for processing the global temporal graph is divided into two parts. The first part comprises three 1D convolutional layers designed to learn the temporal appearance of all the local scenes with boundaries, 
each followed by a sigmoid activation for non-linearity. The second part generates the anchor proposals of the temporally learned features via pre-defined anchors, where each of the anchors acts as a binary mask over the whole graph.

Overall, the GCN module provides two different outputs. i) \textit{Activity classification}: the list of predicted activity labels for all the snippets in the video ($\hat{y}_{{s}_1}$, ..., $\hat{y}_{{s}_i}$, ..., $\hat{y}_{{s}_N}$) is produced, where the dimensionality of the output vector $\hat{y}$ is equal to the number of classes. ii) \textit{Activity localisation}: activities are localised using binary masked class-agnostic anchor proposals. The Intersection over Union (IoU) measure between each anchor and the ground truth (true temporal extension of the activity) is computed, and the anchors with maximum IoU
are selected to train the model for localising the boundaries of any activity, regardless of its activity label. 
The final output of our temporal graph is a one-hot binary vector  ($\hat{y}_{{br}_1}$, ..., $\hat{y}_{{br}_i}$, ..., $\hat{y}_{{br}_N}$) for each series of snippets (video), where 
$\hat{y}_{{br}_i} = 1$ iff snippet $S_i$ belongs to the activity, 
$ = 0$ when the snippet does not belong to it.


\vspace{-2mm}
\section{Experimental Evaluation} \label{sec:experiments}

\vspace{-1mm}
\subsection{Datasets} \label{sec:datasets}
\vspace{-1mm}
\textbf{ROAD} (The ROad event Awareness Dataset for Autonomous Driving) \cite{singh2022road} is a multi-labeled dataset proposed for road agent, action, and location detection. The combination of these three labels is referred to in \cite{singh2022road} as a `road event'. The ROAD dataset consists of a total of 22 videos with an average duration of 8 minutes, captured by the Oxford RobotCar \cite{RobotCarDatasetIJRR} under diverse lighting and weather conditions. The dataset was further extended as a testbed for \cad{} in \cite{skhan2021comp}. 
Complex activities in the ROAD dataset belong to six different classes, including: negotiating an intersection, negotiating a pedestrian crossing, waiting in a queue, merging into the (ego) vehicle lane, sudden appearance (of other vehicles/agents), and (people) walking in the middle of the road. Activities can span up to two minutes and involve a large number of road agents.

\textbf{Thumos-14} \cite{idrees2017thumos} is a benchmark datasets for TAL. It contains 413 untrimmed temporally annotated videos categorised into 20 actions. Videos are characterised by a large variance in duration, from one second to 26 minutes. On average, each video contains 16 action instances. To compare our performance with the state-of-the-art, we adopt standard practice of using the validation set (200 videos) for training while evaluating our model on the test set (213 videos).

\textbf{ActivityNet-1.3} \cite{caba2015activitynet} is one of the largest action localisation datasets with around 20K untrimmed videos comprising 200 action categories. The videos are divided into training, validation, and testing folds according to a ratio of 2:1:1, respectively. The number of action instances per video is 1.65, which is quite low compared to Thumos-14. Following the previous art, we train our model on the training set and test it on the validation set.

\vspace{-1mm}
\subsection{Implementation Details} \label{sec:implem}
\vspace{-1mm}
\textbf{Evaluation metrics}. In our experiments, mean Average Precision (mAP) was used as an evaluation metric, using different IoU thresholds for the different datasets. According to the official protocols for the various benchmarks, the following lists of temporal IoU thresholds were selected: $\{ 0.1,0.2,0.3,0.4,0.5 \}$ for ROAD, $\{ 0.3,0.4,0.5,0.6,0.7 \}$ for Thumos-14 and $\{ 0.5, 0.7, 0.95 \}$ for ActivityNet-1.3.

\textbf{Feature extraction}. Firstly, the agent tubes are constructed by detecting scene objects using a YOLOv5 detector \cite{glenn_jocher_2022_6222936} pre-trained on the COCO dataset. Detections are then tracked throughout a snippet using DeepSort \cite{yolov5deepsort2020}. Then, features are extracted using an I3D network pre-trained on the Kinetics dataset \cite{kay2017kinetics}, from both the entire snippet and each cropped agent tube individually. The object categories were reduced to six for the ROAD dataset to only cover the agents actually present in the road scenes portrayed there. As the other two datasets (ActivityNet-1.3 and Thumos-14) are general purpose, in their case we retained all the 80 classes present in the COCO dataset.

\textbf{Scene Graph Attention}. The local scene graph was generated by producing a list of tuples $[(0, 1), (0, 2), ...]$, where the first index relates to the source node
and the second number indexes the target node. The reason for preferring this structure over an adjacency matrix was to limit memory usage. For node attention learning, {we initialised our architecture using the weights of the GAT model \cite{velivckovic2017graph} pre-trained on the PPI dataset \cite{zitnik2017predicting} and} applied 4 attention layers with \{4, 4, \emph{num of classes}, and \emph{num of classes}\} heads, respectively. The number of classes 
is equal to 201, 21, and 7 in ActivityNet-1.3, Thumos-14 and ROAD, respectively.

The 
\textbf{Temporal Graph} is a stack of three 1D convolution layers on the final representation of the temporally connected local scenes. The size of the input to the first convolutional layer is the number $N$ of local scenes (snippets), 
multiplied by the number of heads in the last attention layer.

The length of our temporal graph is fixed to $N$. Videos with number of snippets less than or equal to $N$ are passed directly to the temporal graph; 
longer videos
are split into multiple chunks containing $N$ snippets each.
The output is a one-hot vector of activity labels of size $N$ and 
a collection of 128 proposals (binary graph masks) also of length $N$.

\textbf{Loss Functions}. Our problem is multi-objective, as we aim at not only recognising the label of the activity taking place but also finding its boundary (start and end time). Our overall loss function is thus the weighted sum of \emph{BCEWithLogitsLoss} \cite{BCEWithLogitsLoss} (for activity classification) and standard binary cross entropy (for temporal localisation). Full details 
can be found in the \textbf{Supplementary material}.

\setlength{\tabcolsep}{4pt}
\begin{table}
\begin{center}
\caption{Comparing our approach with the state-of-the-art methods for \cad{} available on the ROAD dataset. The mAP at the various standard thresholds is reported. Best results are in \textbf{bold} and second best \underline{underlined}.}
\vspace{-2mm}

\vspace{-2mm}
\label{table:road}
{ \begin{tabular}{lllllll}
\hline\noalign{\smallskip}
 &  \multicolumn{6}{c}{ROAD}\\
Methods $\qquad\qquad$ & 0.1 & 0.2 & 0.3 & 0.4 & 0.5 & Avg\\
\noalign{\smallskip}
\hline
\noalign{\smallskip}
P-GCN \cite{zeng2019graph}  &60.0 & 56.7 & 53.9 & 50.5 & 46.4& 53.5\\
G-TAD \cite{xu2020g}   &62.1 & 59.8 & 55.6 & 52.2 & 48.7& 55.6\\
STDSG \cite{skhan2021comp} &  77.3 & 74.6 & 71.2 & \underline{66.7} & \underline{59.4} & \underline{69.8} \\
TallFormer \cite{cheng2022tallformer} & \underline{78.4}	& \underline{74.9}	& 70.3	& 63.8	&57.1&	68.9 \\
ActionFormer \cite{zhang2022actionformer} & 76.5	& 73.7	& \underline{72.6} &	64.4	&58.2&	69.0 \\

\textbf{Ours} & \textbf{82.1} & \textbf{77.4} & \textbf{73.3} & \textbf{69.5} & \textbf{62.9} & \textbf{73.0}\\
\hline
\end{tabular}}
\end{center}
\vspace{-9mm}
\end{table}

\setlength{\tabcolsep}{4pt}
\begin{table*}
\begin{center}
\caption{{Activity detection performance comparison 
on Thumos-14 and ActivityNet-1.3. mAP values (\%) at different IoU thresholds are reported for the test and validation sets of Thumos-14 and ActivityNet-1.3, respectively. The models are grouped by whether the model relies on optical flow (OF) or not. Best results are in \textbf{bold} and second best \underline{underlined}.
}}
\vspace{-3mm}
\label{table:thum_act}
{\begin{tabular}{l|l|l|lllll|l|lll|l}

\hline\noalign{\smallskip}
 & & &\multicolumn{6}{c}{Thumos-14} & \multicolumn{4}{c}{ActivityNet-1.3}\\
Methods & Venue & OF & 0.3 & 0.4 & 0.5 & 0.6 & 0.7 & Average  & 0.5 & 0.75 & 0.95 & Average\\
\noalign{\smallskip}
\hline
\noalign{\smallskip}
BSN \cite{lin2018bsn} &ECCV'18 & \cmark &53.5 & 45.0 &36.9 &28.4 &20.0 &36.8& 46.4& 30.0& 8.0& 30.0\\

P-GCN \cite{zeng2019graph} & ICCV'19 & \cmark & 63.6& 57.8& 49.1& —& —& —& 48.3& 33.2& 3.3& 31.1\\
BMN \cite{lin2019bmn} &ICCV'19 & \cmark & 56.0& 47.4& 38.8& 29.7& 20.5& 38.5& 50.1& 34.8& 8.3& 33.8\\

G-TAD \cite{xu2020g} & CVPR'20 & \cmark & 54.5& 47.6& 40.2& 30.8& 23.4& 39.3& 50.4& 34.6& \underline{9.0}& 34.1\\
BC-GNN \cite{bai2020boundary} & ECCV'20 & \cmark & 57.1& 49.1& 40.4& 31.2& 23.1& 40.2& 50.6& 34.8& \textbf{9.4}& 34.3\\

BSN++ \cite{su2021bsn++} & AAAI'21 & \cmark & 59.9& 49.5& 41.3& 31.9& 22.8& 41.1& 51.3& 35.7& 8.3& 34.9\\
MUSES \cite{liu2021multi} & CVPR'21 & \cmark & 68.9& 64.0& 56.9& 46.3& 31.0& 53.4& 50.0& 35.0& 6.6& 34.0\\
ContextLoc \cite{zhu2021enriching} & ICCV'21 & \cmark & 68.3& 63.8& 54.3& 41.8& 26.2& 50.9& 56.0& 35.2& 3.5 & 34.2\\
CPN \cite{hsieh2022contextual} & WACV'22 & \cmark & 68.2& 62.1& 54.1& 41.5& 28.0&50.7& — &—& —& —\\

RefactorNet \cite{xia2022learning} & CVPR'22 & \cmark & 70.7& 65.4& 58.6 & 47.0& 32.1& 54.8& \underline{56.6} & \textbf{40.7}& 7.4& \textbf{38.6}\\
RCL \cite{wang2022rcl} & CVPR'22 & \cmark & 70.1 &62.3& 52.9& 42.7& 30.7& 51.7& 55.1& \underline{39.0} & 8.3& \underline{37.6}\\
LDCLR \cite{zhu2022learning} & AAAI'22 & \cmark & 72.1 & 65.9& 57.0& 44.2& 28.5& 53.5& \textbf{58.1} & 36.3& 6.2& 35.2\\

ActionFormer \cite{zhang2022actionformer} & ECCV'22 & \cmark & \underline{82.1} & \underline{77.8} & \underline{71.0} & \underline{59.4} & \underline{43.9} & \underline{66.8} & 53.5 & 36.2 & 8.2 & 35.6 \\

Re$^2$TAL \cite{zhao2023re2tal} & CVPR'23 & \cmark & 77.4 & 72.6 & 64.9 & 53.7 & 39.0 & 61.5
& 55.3 & 37.9 & 9.0& 37.0 \\

TriDet \cite{shi2023tridet} &CVPR'23 & \cmark & \textbf{83.6} & \textbf{80.1} & \textbf{72.9} & \textbf{62.4} & \textbf{47.4} & \textbf{69.3} & 54.7 & 38.0 & 8.4 & 36.8 \\

\hline
GTAN \cite{long2019gaussian} & CVPR'19 & \xmark & 57.8& 47.2& 38.8& — &— &— &52.6 &34.1& 8.9& 34.3\\

TadTR \cite{liu2022end} & TIP’22 & \xmark  & 59.6 &  54.5 & 47.0 & 37.8 & 26.5 & 45.1 & 49.6 & 35.2 & 9.9 & 34.3 \\
 
E2E-TAD \cite{liu2022empirical} & CVPR'22 & \xmark & 69.4& 64.3& 56.0& 46.4& 34.9& 54.2& 50.8& 36.0 & \underline{10.8} & 35.1\\
ActionFormer \cite{zhang2022actionformer} & ECCV'22 & \xmark & 69.8 & \underline{66.0} & 58.7 & 48.3 & 34.6 & 55.5 & 53.2 & 35.1 & 8.0 & 34.9 \\

TAGS \cite{nag2022proposal} & ECCV’22 & \xmark & 59.8 & 57.2 & 50.7 & 42.6 & 29.1 & 47.9 & \underline{54.4} & 34.9 & 8.7 & 34.9\\

TallFormer \cite{cheng2022tallformer} & ECCV'22 & \xmark & \underline{76.1} &— & \textbf{63.2} &— & 34.5 & \underline{59.2} & 54.1 & \underline{36.2} & 7.9 & 35.5 \\

DL-Net \cite{you2023dl} & ICASSP'23 & \xmark & 61.3 & 55.8 & 47.7 & 37.6 & 26.4& --
& 50.3& 35.0& 9.3 &34.3
 \\
DCMD \cite{lee2023decomposed} & CVPR'23 & \xmark & 70.5 & 65.8 &59.2 & \textbf{50.1}& \textbf{38.2} & 56.8
& 53.7& 35.9& 8.6 & \underline{35.6} \\

\textbf{Ours} & — & \xmark & \textbf{78.2} & \textbf{69.5} & \underline{62.7} & \textbf{50.1} & \underline{36.9} & \textbf{59.8} & \textbf{60.6} & \textbf{40.3} & \textbf{11.1} & \textbf{39.3}\\

\hline
\end{tabular}}
\end{center}
\vspace{-9mm}
\end{table*}
\vspace{-2mm}
\subsection{Comparison with State-of-the-Art}
\vspace{-1mm}
{
\textbf{ROAD}. To validate our \cad{} approach, we first compared ourselves with the prior art available for the ROAD dataset \emph{and} re-implemented the current state-of-the-art TAL methods; TallFormer \cite{cheng2022tallformer} and ActionFormer \cite{zhang2022actionformer} there 
(see Table \ref{table:road}). The standard IoU thresholds for mAP computation on ROAD range from 0.1 to 0.5.


\textbf{Thumos-14}. Our hybrid graph approach is compared with the state of the art on Thumos-14 in Table \ref{table:thum_act}. 
We followed the standard evaluation metric there: mAP with IoU thresholds ranging from 0.3 to 0.7 and their average. 
Competitor methods were published in top computer vision
venus in 2018-2023. These methods are grouped into two categories whether the optical flow (OF) modality is used or not especially at inference time.
The recent TriDet \cite{shi2023tridet} approach achieves the highest average mAP 69.3 by using both OF and RGB modalities while our method outperforms all RGB-based models. 
\begin{figure}[h!]
    \centering
    \vspace{-2mm}
    \includegraphics[width=0.46\textwidth]{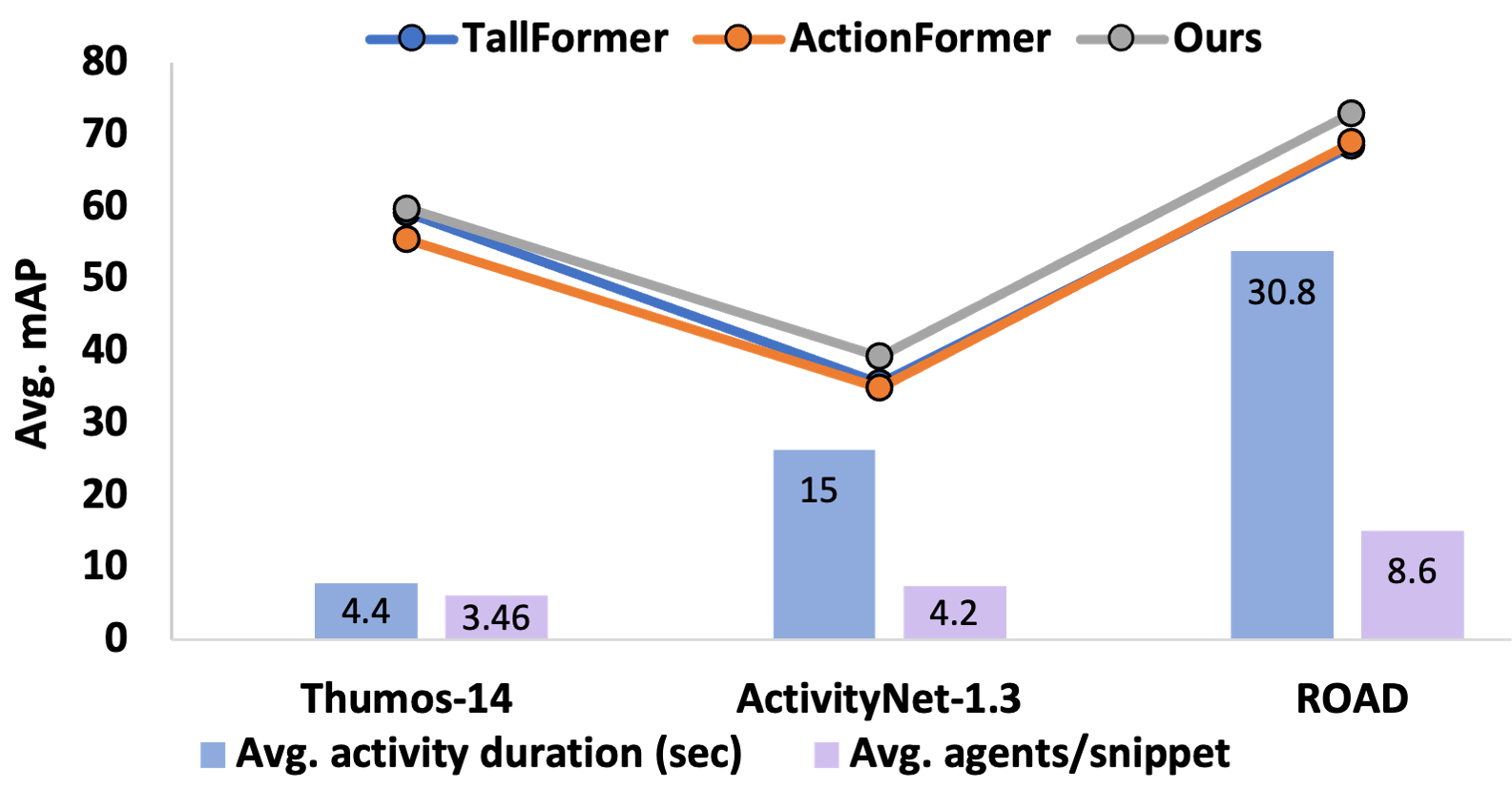}
    \vspace{-3mm}
    \caption{{Performance comparison with the state-of-the-art over all the three datasets considered, as a function of 
    average activity duration and average no. of agents per snippet. \label{fig:sota}}}
    \vspace{-4mm}
\end{figure}

\textbf{ActivityNet-1.3}. Table \ref{table:thum_act} also compares our proposal with the state-of-the-art methods on ActivityNet-1.3. The mAP IoU thresholds used for comparison are 0.5, 0.75, and 0.95. 
Our approach achieves the best mAP for all IoU thresholds and even outperforms all the methods including the one using both OF and RGB modalities.

To sum up, our proposal clearly outperforms all previous approaches over the ROAD dataset, showing the potential of this approach for modelling and detecting long, complex activities. Further, our approach outperforms all competitors over Thumos-14 and ActivityNet.
The comparison of methods with respect to the complexity of the datasets is illustrated in 
Fig. \ref{fig:sota} shows how increasingly outperforms the prior art as the duration and complexity of the activities increases.
}

\vspace{-2mm}
\subsection{Ablation Studies}\label{sec:ablation}

\begin{figure}[h!]
    \centering
\includegraphics[width=0.39\textwidth]{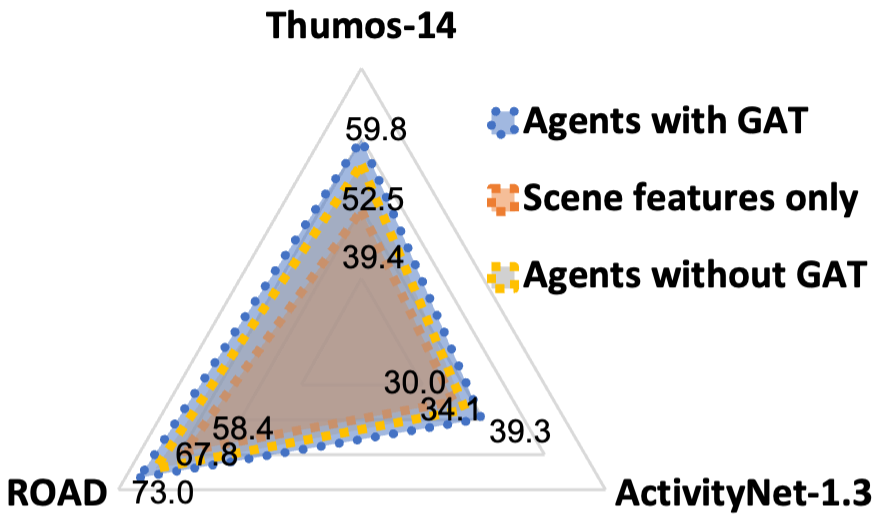}
    \caption{{Average mAP of variants of our method with: attended scene graph ('Agents with GAT'), scene graph ('Agents without GAT') and 'Scene features only', over the all three datasets. 
    }    }
    \vspace{-7mm}
    \label{fig:agent_nodes}
\end{figure}

\vspace{-1mm}
\textbf{Effect of Agent Nodes}. Firstly, we showed the advantage of 
{ modelling the scene as a graph of agents,}
compared to simply using whole scene features. Namely, we removed the local scene graph from our pipeline (Fig. \ref{fig:framework}, stage B) and passed the 3D features of the whole scene as a node to the global temporal graph. The significant performance drop can be clearly observed in Fig. \ref{fig:agent_nodes} over all three datasets.

\textbf{Effect of Edges and Aggregation}.
{
Next, we studied the influence of different types of edge connections in the local scene graph: (i)
a fully-connected scene graph;
(ii) a star structure where each agent node is connected to the scene node only; 
(iii) a star structure with additional connections between agent nodes sharing the same label.}

We also validated two different techniques for extracting the final representation from the local scene graph, named `Aggregated' and `Scene'. In the former, the feature representation is extracted by aggregating 
those of all the attended nodes. In the latter, only the feature vector related to the scene node (after attention) is retained. 
Table \ref{table:edge_agg} shows the effect of all possible combinations of graph topologies and aggregation strategies. A fully-connected scene graph with Aggregated features performs best in two cases over three, while the star topology best suits ActivityNet. 

\begin{table}[ht!]
\vspace{-1mm}
\caption{{Effect of local scene graph topologies and feature aggregation strategies on the performance of our proposal. 
\vspace{-3mm}
\label{table:edge_agg}
}}

{\begin{tabular}{lll|ll|ll}
\hline\noalign{\smallskip}
$\qquad\qquad$ &  \multicolumn{2}{c}{Thumos-14} & \multicolumn{2}{c}{Act.Net-1.3} & \multicolumn{2}{c}{ROAD}\\
Topology & Aggr. & Scene & Aggr. & Scene & Aggr. & Scene\\
\noalign{\smallskip}
\hline
\noalign{\smallskip}
Fully & \textbf{59.8} & 49.2 & 37.4 & 31.4 & \textbf{73.0} & 62.7\\
Star & 51.2 & 44.8 & \textbf{39.3} & 36.6 & 62.3 & 57.9\\
Star + & 52.2 & 41.9 & 35.3 &32.7 &64.9   & 59.2\\
\hline
\end{tabular}}
\vspace{-1mm}
\end{table}

\textbf{Effect of Sequence Length}. To explore the effect on our model of snippet duration (the temporal extent of the local dynamic scene), 
we performed experiments with four different sizes (12, 18, 24, and 30), reported in Table \ref{table:seq_lens}. On ActivityNet-1.3 and ROAD, the top scores were obtained by selecting a sequence length of 24, due to the nature of the activities present in these datasets, which last longer. On the other hand, on Thumos-14 we achieved the best performance using a sequence length of 18, as most activities there are shorter in duration.

\begin{table} [h!]
\centering
\begin{center}
\vspace{-1mm}
\caption{{Average mAP over all IoU thresholds of our method as a function of different snippet lengths for the three datasets. 
}}
\vspace{-3mm}

\label{table:seq_lens}
{\begin{tabular}{ll|l|l}
\hline\noalign{\smallskip}
Snippet length & Thumos-14 & ActivityNet-1.3 & ROAD\\
\noalign{\smallskip}
\hline
\noalign{\smallskip}
12 & 52.7 & 31.3 & 56.9 \\
18 & \textbf{59.8} & 37.6 & 62.6 \\
24 & 54.3 & \textbf{39.3} & \textbf{73.0} \\
30 & 49.7 & 34.5 & 70.0 \\
\hline
\end{tabular}}
\end{center}
\vspace{-5mm}
\end{table}

\textbf{Effect of Temporal Graph Length}. 
We also ablated the effect of the length of the temporal graph, i.e., 
the number of local scene nodes composing the global graph (Table \ref{table:temp_len}). 
Four different graph lengths $(128, 256, 512, 1024)$ are compared. The best performance is achieved on Thumos-14  using a smaller number of scene nodes, due to the average duration of the videos there and its portraying shorter-term activities. In contrast, on ROAD and ActivityNet-1.3 our approach performed the best using longer temporal graphs. 

\begin{table}
\centering
\begin{center}
\vspace{-2mm}
\caption{{
Average mAP over all IoU thresholds of our method as a function of different temporal graph lengths for the three datasets.
}}
\vspace{-3mm}
\label{table:temp_len}
{ \begin{tabular}{ll|l|l}
\hline\noalign{\smallskip}
Temporal length & Thumos-14 & ActivityNet-1.3 & ROAD\\
\noalign{\smallskip}
\hline
\noalign{\smallskip}
128 & \textbf{59.8} & 29.6 & 48.3 \\
256 & 52.9 & 32.7 & 58.5 \\
512 & 50.3 & \textbf{39.3} & 69.2  \\
1,024 & 46.8 & 34.1 & \textbf{73.0} \\
\hline
\end{tabular}}
\end{center}
\vspace{-3mm}
\end{table}


\begin{figure}[h]
    \centering
    \includegraphics[width=0.42\textwidth]{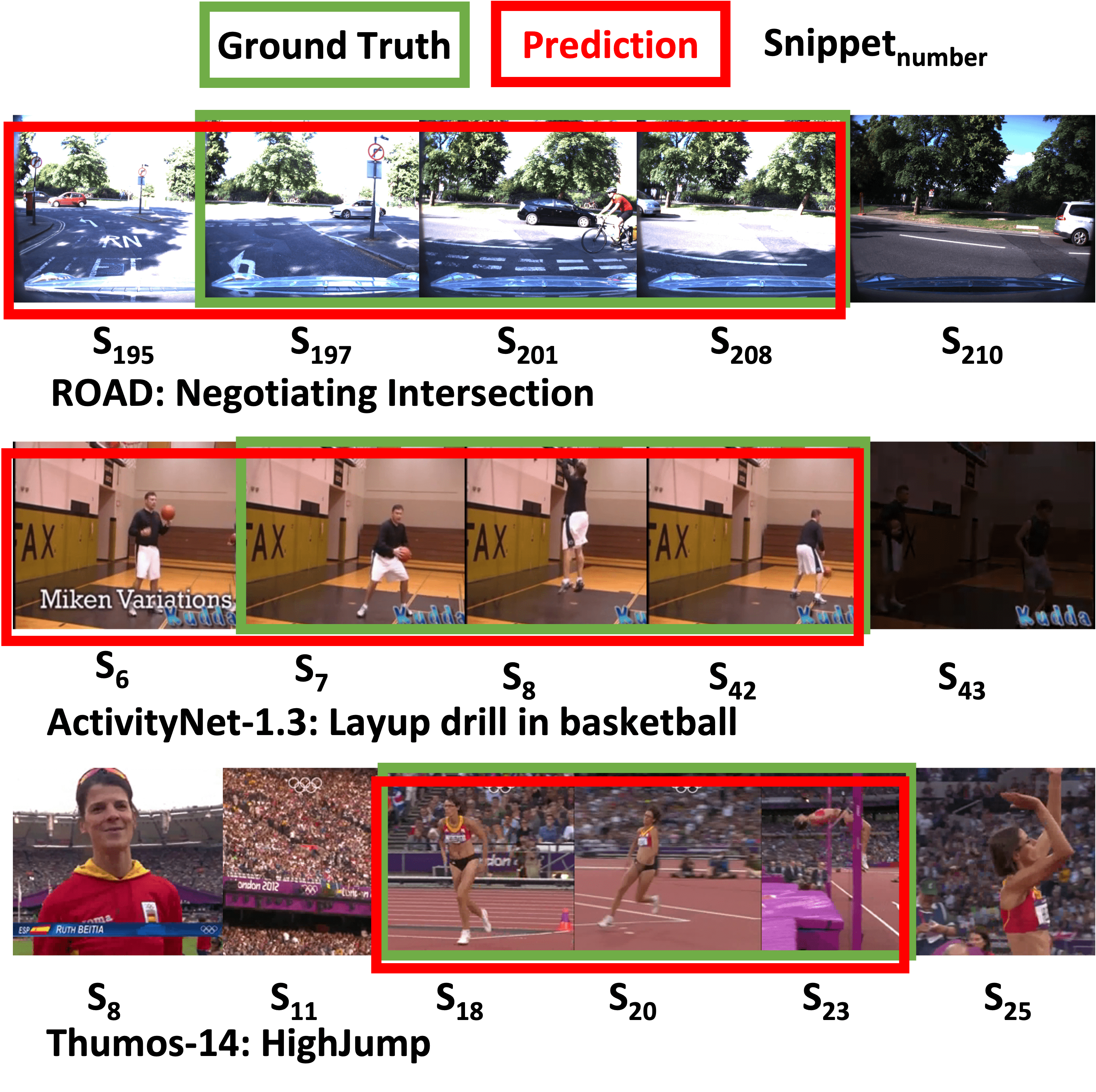}
    \vspace{-4mm}
    \caption{Qualitative results of our method on all 3 datasets. The green rectangles spanning the snippets (local scenes) are the ground truth; the red boxes denote our model's predictions.}
    \label{fig:qaul_res}
    \vspace{-5mm}
\end{figure}

\textbf{Qualitative Results}. 
To help the reader visualise the output of our proposed method, we show some qualitative detection results on all three datasets in Fig. \ref{fig:qaul_res}. The figure shows one sample per dataset, and portrays a series of local scenes (snippets), skipping some for visualisation purposes, with superimposed the ground truth (in green) and the prediction of our model (in red). For example, for ActivityNet-1.3 an instance of the `Layup drill in basketball' class is shown in which the activity starts with snippet 7 and ends with snippet 42. Our model predicts the activity to start from snippet 6 and end with snippet 43.


\vspace{-2mm}
\section{Conclusions}
\vspace{-1mm}
This paper explicitly addresses the problem of detecting longer-term, complex activities using
a novel hybrid graph neural network-based 
framework - combining both scene graph attention and a temporal graph to model activities of arbitrary duration. Our proposed framework is divided into three main building blocks: agent tube detection and feature extraction; a local scene graph construction with attention; 
and a temporal graph for recognising the class label and localising each activity instance.
We tested our method on three benchmark datasets, showing the effectiveness of our method in detecting both short-term and long-term activities, thanks to its ability to model their finer-grained structure without the need for extra annotation. Our approach outperforms all previous state-of-the-art methods on all of the datasets including Thumos-14, ActivityNet-1.3, and ROAD datasets.

In the future, we intend to progress from incremental inference to incremental training, by learning to construct activity graphs in an incremental manner, paving the way to applications such as future activity anticipation \cite{liu2022hybrid} and pedestrian intent prediction \cite{cadena2022pedestrian}. 
A further exciting line of research is the modelling of the uncertainty associated with complex scenes, in either the Bayesian \cite{kendall2017uncertainties} 
or the full epistemic settings \cite{manchingal2022epistemic,osband2021epistemic}.

\section*{Acknowledgements}
This project has received funding from the European Union’s Horizon 2020 research and innovation programme, under grant agreement No. 964505 (E-pi).

{\small
\bibliographystyle{ieee_fullname}
\bibliography{egbib}

\begin{thebibliography}{10}\itemsep=-1pt

\bibitem{BCEWithLogitsLoss}
Bcewithlogitsloss function.
\newblock
  \url{https://pytorch.org/docs/stable/generated/torch.nn.BCEWithLogitsLoss.html}.
\newblock Accessed: 2022-11-01.

\bibitem{bai2020boundary}
Yueran Bai, Yingying Wang, Yunhai Tong, Yang Yang, Qiyue Liu, and Junhui Liu.
\newblock Boundary content graph neural network for temporal action proposal
  generation.
\newblock In {\em European Conference on Computer Vision}, pages 121--137.
  Springer, 2020.

\bibitem{bao2022opental}
Wentao Bao, Qi Yu, and Yu Kong.
\newblock Opental: Towards open set temporal action localization.
\newblock In {\em Proceedings of the IEEE/CVF Conference on Computer Vision and
  Pattern Recognition}, pages 2979--2989, 2022.

\bibitem{yolov5deepsort2020}
Mikel Broström.
\newblock Real-time multi-object tracker using yolov5 and deep sort.
\newblock \url{https://github.com/mikel-brostrom/Yolov5_DeepSort_Pytorch},
  2020.

\bibitem{buch2019end}
Shyamal Buch, Victor Escorcia, Bernard Ghanem, Li Fei-Fei, and Juan~Carlos
  Niebles.
\newblock End-to-end, single-stream temporal action detection in untrimmed
  videos.
\newblock 2019.

\bibitem{Buchner20223DMT}
Martin Buchner and Abhinav Valada.
\newblock 3d multi-object tracking using graph neural networks with cross-edge
  modality attention.
\newblock {\em ArXiv}, abs/2203.10926, 2022.

\bibitem{caba2015activitynet}
Fabian Caba~Heilbron, Victor Escorcia, Bernard Ghanem, and Juan Carlos~Niebles.
\newblock Activitynet: A large-scale video benchmark for human activity
  understanding.
\newblock In {\em Proceedings of the ieee conference on computer vision and
  pattern recognition}, pages 961--970, 2015.

\bibitem{cadena2022pedestrian}
Pablo Rodrigo~Gantier Cadena, Yeqiang Qian, Chunxiang Wang, and Ming Yang.
\newblock Pedestrian graph+: A fast pedestrian crossing prediction model based
  on graph convolutional networks.
\newblock {\em IEEE Transactions on Intelligent Transportation Systems}, 2022.

\bibitem{caesar2020nuscenes}
Holger Caesar, Varun Bankiti, Alex~H Lang, Sourabh Vora, Venice~Erin Liong,
  Qiang Xu, Anush Krishnan, Yu Pan, Giancarlo Baldan, and Oscar Beijbom.
\newblock nuscenes: A multimodal dataset for autonomous driving.
\newblock In {\em Proceedings of the IEEE/CVF conference on computer vision and
  pattern recognition}, pages 11621--11631, 2020.

\bibitem{carreira2017quo}
Joao Carreira and Andrew Zisserman.
\newblock Quo vadis, action recognition? a new model and the kinetics dataset.
\newblock In {\em proceedings of the IEEE Conference on Computer Vision and
  Pattern Recognition}, pages 6299--6308, 2017.

\bibitem{chao2018rethinking}
Yu-Wei Chao, Sudheendra Vijayanarasimhan, Bryan Seybold, David~A Ross, Jia
  Deng, and Rahul Sukthankar.
\newblock Rethinking the faster r-cnn architecture for temporal action
  localization.
\newblock In {\em proceedings of the IEEE conference on computer vision and
  pattern recognition}, pages 1130--1139, 2018.

\bibitem{cheng2022tallformer}
Feng Cheng and Gedas Bertasius.
\newblock Tallformer: Temporal action localization with long-memory
  transformer.
\newblock {\em arXiv preprint arXiv:2204.01680}, 2022.

\bibitem{hsieh2022contextual}
He-Yen Hsieh, Ding-Jie Chen, and Tyng-Luh Liu.
\newblock Contextual proposal network for action localization.
\newblock In {\em Proceedings of the IEEE/CVF Winter Conference on Applications
  of Computer Vision}, pages 2129--2138, 2022.

\bibitem{hu2020progressive}
Guyue Hu, Bo Cui, Yuan He, and Shan Yu.
\newblock Progressive relation learning for group activity recognition.
\newblock In {\em Proceedings of the IEEE/CVF Conference on Computer Vision and
  Pattern Recognition}, pages 980--989, 2020.

\bibitem{huang2019decoupling}
Yupan Huang, Qi Dai, and Yutong Lu.
\newblock Decoupling localization and classification in single shot temporal
  action detection.
\newblock In {\em 2019 IEEE International Conference on Multimedia and Expo
  (ICME)}, pages 1288--1293. IEEE, 2019.

\bibitem{idrees2017thumos}
Haroon Idrees, Amir~R Zamir, Yu-Gang Jiang, Alex Gorban, Ivan Laptev, Rahul
  Sukthankar, and Mubarak Shah.
\newblock The thumos challenge on action recognition for videos “in the
  wild”.
\newblock {\em Computer Vision and Image Understanding}, 155:1--23, 2017.

\bibitem{glenn_jocher_2022_6222936}
Glenn Jocher, Ayush Chaurasia, Alex Stoken, Jirka Borovec, NanoCode012, Yonghye
  Kwon, TaoXie, Jiacong Fang, imyhxy, Kalen Michael, Lorna, Abhiram V, Diego
  Montes, Jebastin Nadar, Laughing, tkianai, yxNONG, Piotr Skalski, Zhiqiang
  Wang, Adam Hogan, Cristi Fati, Lorenzo Mammana, AlexWang1900, Deep Patel,
  Ding Yiwei, Felix You, Jan Hajek, Laurentiu Diaconu, and Mai~Thanh Minh.
\newblock {ultralytics/yolov5: v6.1 - TensorRT, TensorFlow Edge TPU and
  OpenVINO Export and Inference}, Feb. 2022.

\bibitem{kay2017kinetics}
Will Kay, Joao Carreira, Karen Simonyan, Brian Zhang, Chloe Hillier, Sudheendra
  Vijayanarasimhan, Fabio Viola, Tim Green, Trevor Back, Paul Natsev, et~al.
\newblock The kinetics human action video dataset.
\newblock {\em arXiv preprint arXiv:1705.06950}, 2017.

\bibitem{kendall2017uncertainties}
Alex Kendall and Yarin Gal.
\newblock What uncertainties do we need in bayesian deep learning for computer
  vision?
\newblock {\em Advances in neural information processing systems}, 30, 2017.

\bibitem{khan2021spatiotemporal}
Salman Khan and Fabio Cuzzolin.
\newblock Spatiotemporal deformable models for long-term complex activity
  detection.
\newblock {\em arXiv preprint arXiv:2104.08194}, 2021.

\bibitem{skhan2021comp}
Salman Khan and Fabio Cuzzolin.
\newblock Spatiotemporal deformable scene graphs for complex activity
  detection.
\newblock In {\em 32nd British Machine Vision Conference 2021, {BMVC} 2021,
  Online, November 22-25}, 2021.

\bibitem{kosaraju2019social}
Vineet Kosaraju, Amir Sadeghian, Roberto Mart{\'\i}n-Mart{\'\i}n, Ian Reid,
  Hamid Rezatofighi, and Silvio Savarese.
\newblock Social-bigat: Multimodal trajectory forecasting using bicycle-gan and
  graph attention networks.
\newblock {\em Advances in Neural Information Processing Systems}, 32, 2019.

\bibitem{krishna2017dense}
Ranjay Krishna, Kenji Hata, Frederic Ren, Li Fei-Fei, and Juan Carlos~Niebles.
\newblock Dense-captioning events in videos.
\newblock In {\em Proceedings of the IEEE international conference on computer
  vision}, pages 706--715, 2017.

\bibitem{lee2023decomposed}
Pilhyeon Lee, Taeoh Kim, Minho Shim, Dongyoon Wee, and Hyeran Byun.
\newblock Decomposed cross-modal distillation for rgb-based temporal action
  detection.
\newblock In {\em Proceedings of the IEEE/CVF Conference on Computer Vision and
  Pattern Recognition}, pages 2373--2383, 2023.

\bibitem{li2021multisports}
Yixuan Li, Lei Chen, Runyu He, Zhenzhi Wang, Gangshan Wu, and Limin Wang.
\newblock Multisports: A multi-person video dataset of spatio-temporally
  localized sports actions.
\newblock In {\em Proceedings of the IEEE/CVF International Conference on
  Computer Vision}, pages 13536--13545, 2021.

\bibitem{lin2021learning}
Chuming Lin, Chengming Xu, Donghao Luo, Yabiao Wang, Ying Tai, Chengjie Wang,
  Jilin Li, Feiyue Huang, and Yanwei Fu.
\newblock Learning salient boundary feature for anchor-free temporal action
  localization.
\newblock In {\em Proceedings of the IEEE/CVF Conference on Computer Vision and
  Pattern Recognition}, pages 3320--3329, 2021.

\bibitem{lin2019bmn}
Tianwei Lin, Xiao Liu, Xin Li, Errui Ding, and Shilei Wen.
\newblock Bmn: Boundary-matching network for temporal action proposal
  generation.
\newblock In {\em Proceedings of the IEEE/CVF international conference on
  computer vision}, pages 3889--3898, 2019.

\bibitem{lin2018bsn}
Tianwei Lin, Xu Zhao, Haisheng Su, Chongjing Wang, and Ming Yang.
\newblock Bsn: Boundary sensitive network for temporal action proposal
  generation.
\newblock In {\em Proceedings of the European conference on computer vision
  (ECCV)}, pages 3--19, 2018.

\bibitem{lin2014microsoft}
Tsung-Yi Lin, Michael Maire, Serge Belongie, James Hays, Pietro Perona, Deva
  Ramanan, Piotr Doll{\'a}r, and C~Lawrence Zitnick.
\newblock Microsoft coco: Common objects in context.
\newblock In {\em European conference on computer vision}, pages 740--755.
  Springer, 2014.

\bibitem{liu2022hybrid}
Tianshan Liu and Kin-Man Lam.
\newblock A hybrid egocentric activity anticipation framework via
  memory-augmented recurrent and one-shot representation forecasting.
\newblock In {\em Proceedings of the IEEE/CVF Conference on Computer Vision and
  Pattern Recognition}, pages 13904--13913, 2022.

\bibitem{liu2022empirical}
Xiaolong Liu, Song Bai, and Xiang Bai.
\newblock An empirical study of end-to-end temporal action detection.
\newblock In {\em Proceedings of the IEEE/CVF Conference on Computer Vision and
  Pattern Recognition}, pages 20010--20019, 2022.

\bibitem{liu2021multi}
Xiaolong Liu, Yao Hu, Song Bai, Fei Ding, Xiang Bai, and Philip~HS Torr.
\newblock Multi-shot temporal event localization: a benchmark.
\newblock In {\em Proceedings of the IEEE/CVF Conference on Computer Vision and
  Pattern Recognition}, pages 12596--12606, 2021.

\bibitem{liu2022end}
Xiaolong Liu, Qimeng Wang, Yao Hu, Xu Tang, Shiwei Zhang, Song Bai, and Xiang
  Bai.
\newblock End-to-end temporal action detection with transformer.
\newblock {\em IEEE Transactions on Image Processing}, 31:5427--5441, 2022.

\bibitem{liu2019multi}
Yuan Liu, Lin Ma, Yifeng Zhang, Wei Liu, and Shih-Fu Chang.
\newblock Multi-granularity generator for temporal action proposal.
\newblock In {\em Proceedings of the IEEE/CVF conference on computer vision and
  pattern recognition}, pages 3604--3613, 2019.

\bibitem{long2019gaussian}
Fuchen Long, Ting Yao, Zhaofan Qiu, Xinmei Tian, Jiebo Luo, and Tao Mei.
\newblock Gaussian temporal awareness networks for action localization.
\newblock In {\em Proceedings of the IEEE/CVF Conference on Computer Vision and
  Pattern Recognition}, pages 344--353, 2019.

\bibitem{RobotCarDatasetIJRR}
Will Maddern, Geoff Pascoe, Chris Linegar, and Paul Newman.
\newblock {1 Year, 1000km: The Oxford RobotCar Dataset}.
\newblock {\em The International Journal of Robotics Research (IJRR)},
  36(1):3--15, 2017.

\bibitem{manchingal2022epistemic}
Shireen~Kudukkil Manchingal and Fabio Cuzzolin.
\newblock Epistemic deep learning.
\newblock {\em arXiv preprint arXiv:2206.07609}, 2022.

\bibitem{mun2019streamlined}
Jonghwan Mun, Linjie Yang, Zhou Ren, Ning Xu, and Bohyung Han.
\newblock Streamlined dense video captioning.
\newblock In {\em Proceedings of the IEEE/CVF conference on computer vision and
  pattern recognition}, pages 6588--6597, 2019.

\bibitem{nag2022proposal}
Sauradip Nag, Xiatian Zhu, Yi-Zhe Song, and Tao Xiang.
\newblock Proposal-free temporal action detection via global segmentation mask
  learning.
\newblock In {\em European Conference on Computer Vision}, pages 645--662.
  Springer, 2022.

\bibitem{nawhal2021activity}
Megha Nawhal and Greg Mori.
\newblock Activity graph transformer for temporal action localization.
\newblock {\em arXiv preprint arXiv:2101.08540}, 2021.

\bibitem{osband2021epistemic}
Ian Osband, Zheng Wen, Mohammad Asghari, Morteza Ibrahimi, Xiyuan Lu, and
  Benjamin Van~Roy.
\newblock Epistemic neural networks.
\newblock {\em arXiv preprint arXiv:2107.08924}, 2021.

\bibitem{shi2023tridet}
Dingfeng Shi, Yujie Zhong, Qiong Cao, Lin Ma, Jia Li, and Dacheng Tao.
\newblock Tridet: Temporal action detection with relative boundary modeling.
\newblock In {\em Proceedings of the IEEE/CVF Conference on Computer Vision and
  Pattern Recognition}, pages 18857--18866, 2023.

\bibitem{shou2016temporal}
Zheng Shou, Dongang Wang, and Shih-Fu Chang.
\newblock Temporal action localization in untrimmed videos via multi-stage
  cnns.
\newblock In {\em Proceedings of the IEEE conference on computer vision and
  pattern recognition}, pages 1049--1058, 2016.

\bibitem{singh2022road}
Gurkirt Singh, Stephen Akrigg, Manuele Di~Maio, Valentina Fontana, Salman Khan,
  Suman Saha, Kossar Jeddisaravi, Farzad Yousefi, Jacob Culley, Thomas
  Nicholson, et~al.
\newblock Road: The road event awareness dataset for autonomous driving.
\newblock {\em IEEE Transactions on Pattern Analysis and Machine Intelligence},
  2022.

\bibitem{song2019session}
Weiping Song, Zhiping Xiao, Yifan Wang, Laurent Charlin, Ming Zhang, and Jian
  Tang.
\newblock Session-based social recommendation via dynamic graph attention
  networks.
\newblock In {\em Proceedings of the Twelfth ACM international conference on
  web search and data mining}, pages 555--563, 2019.

\bibitem{su2021bsn++}
Haisheng Su, Weihao Gan, Wei Wu, Yu Qiao, and Junjie Yan.
\newblock Bsn++: Complementary boundary regressor with scale-balanced relation
  modeling for temporal action proposal generation.
\newblock In {\em Proceedings of the AAAI Conference on Artificial
  Intelligence}, volume~35, pages 2602--2610, 2021.

\bibitem{izzeddin2022}
Izzeddin Teeti, Salman Khan, Ajmal Shahbaz, Andrew Bradley, and Fabio Cuzzolin.
\newblock Vision-based intention and trajectory prediction in autonomous
  vehicles: A survey.
\newblock In Lud~De Raedt, editor, {\em Proceedings of the Thirty-First
  International Joint Conference on Artificial Intelligence, {IJCAI-22}}, pages
  5630--5637. International Joint Conferences on Artificial Intelligence
  Organization, 7 2022.
\newblock Survey Track.

\bibitem{velivckovic2017graph}
Petar Veli{\v{c}}kovi{\'c}, Guillem Cucurull, Arantxa Casanova, Adriana Romero,
  Pietro Lio, and Yoshua Bengio.
\newblock Graph attention networks.
\newblock {\em arXiv preprint arXiv:1710.10903}, 2017.

\bibitem{wang2022rcl}
Qiang Wang, Yanhao Zhang, Yun Zheng, and Pan Pan.
\newblock Rcl: Recurrent continuous localization for temporal action detection.
\newblock In {\em Proceedings of the IEEE/CVF Conference on Computer Vision and
  Pattern Recognition}, pages 13566--13575, 2022.

\bibitem{wang2021oadtr}
Xiang Wang, Shiwei Zhang, Zhiwu Qing, Yuanjie Shao, Zhengrong Zuo, Changxin
  Gao, and Nong Sang.
\newblock Oadtr: Online action detection with transformers.
\newblock In {\em Proceedings of the IEEE/CVF International Conference on
  Computer Vision}, pages 7565--7575, 2021.

\bibitem{Bao_2022_CVPR}
Yu~Kong Wentao~Bao, Qi~Yu.
\newblock Opental: Towards open set temporal action localization.
\newblock In {\em Proceedings of the IEEE/CVF Conference on Computer Vision and
  Pattern Recognition (CVPR)}, June 2022.

\bibitem{xia2022learning}
Kun Xia, Le Wang, Sanping Zhou, Nanning Zheng, and Wei Tang.
\newblock Learning to refactor action and co-occurrence features for temporal
  action localization.
\newblock In {\em Proceedings of the IEEE/CVF Conference on Computer Vision and
  Pattern Recognition}, pages 13884--13893, 2022.

\bibitem{xia2023nested}
Yanjie Xia, Shaochen Wang, and Zhen Kan.
\newblock A nested u-structure for instrument segmentation in robotic surgery.
\newblock In {\em 2023 International Conference on Advanced Robotics and
  Mechatronics (ICARM)}, pages 994--999. IEEE, 2023.

\bibitem{xiao2020convolutional}
Shuwen Xiao, Zhou Zhao, Zijian Zhang, Xiaohui Yan, and Min Yang.
\newblock Convolutional hierarchical attention network for query-focused video
  summarization.
\newblock In {\em Proceedings of the AAAI Conference on Artificial
  Intelligence}, volume~34, pages 12426--12433, 2020.

\bibitem{xu2020g}
Mengmeng Xu, Chen Zhao, David~S Rojas, Ali Thabet, and Bernard Ghanem.
\newblock G-tad: Sub-graph localization for temporal action detection.
\newblock In {\em Proceedings of the IEEE/CVF Conference on Computer Vision and
  Pattern Recognition}, pages 10156--10165, 2020.

\bibitem{yang2022acgnet}
Zichen Yang, Jie Qin, and Di Huang.
\newblock Acgnet: Action complement graph network for weakly-supervised
  temporal action localization.
\newblock In {\em Proceedings of the AAAI Conference on Artificial
  Intelligence}, volume~36, pages 3090--3098, 2022.

\bibitem{you2023dl}
Dianlong You, Houlin Wang, Bingxin Liu, Yang Yu, and Zhiming Li.
\newblock Dl-net: Dilation location network for temporal action detection.
\newblock In {\em ICASSP 2023-2023 IEEE International Conference on Acoustics,
  Speech and Signal Processing (ICASSP)}, pages 1--5. IEEE, 2023.

\bibitem{yuan2017temporal}
Zehuan Yuan, Jonathan~C Stroud, Tong Lu, and Jia Deng.
\newblock Temporal action localization by structured maximal sums.
\newblock In {\em Proceedings of the IEEE Conference on Computer Vision and
  Pattern Recognition}, pages 3684--3692, 2017.

\bibitem{zeng2019graph}
Runhao Zeng, Wenbing Huang, Mingkui Tan, Yu Rong, Peilin Zhao, Junzhou Huang,
  and Chuang Gan.
\newblock Graph convolutional networks for temporal action localization.
\newblock In {\em Proceedings of the IEEE/CVF International Conference on
  Computer Vision}, pages 7094--7103, 2019.

\bibitem{zeng2021graph}
Runhao Zeng, Wenbing Huang, Mingkui Tan, Yu Rong, Peilin Zhao, Junzhou Huang,
  and Chuang Gan.
\newblock Graph convolutional module for temporal action localization in
  videos.
\newblock {\em IEEE Transactions on Pattern Analysis and Machine Intelligence},
  2021.

\bibitem{zhang2022actionformer}
Chen-Lin Zhang, Jianxin Wu, and Yin Li.
\newblock Actionformer: Localizing moments of actions with transformers.
\newblock In {\em European Conference on Computer Vision}, pages 492--510.
  Springer, 2022.

\bibitem{zhao2023re2tal}
Chen Zhao, Shuming Liu, Karttikeya Mangalam, and Bernard Ghanem.
\newblock Re2tal: Rewiring pretrained video backbones for reversible temporal
  action localization.
\newblock In {\em Proceedings of the IEEE/CVF Conference on Computer Vision and
  Pattern Recognition}, pages 10637--10647, 2023.

\bibitem{zhu2021enriching}
Zixin Zhu, Wei Tang, Le Wang, Nanning Zheng, and Gang Hua.
\newblock Enriching local and global contexts for temporal action localization.
\newblock In {\em Proceedings of the IEEE/CVF International Conference on
  Computer Vision}, pages 13516--13525, 2021.

\bibitem{zhu2022learning}
Zixin Zhu, Le Wang, Wei Tang, Ziyi Liu, Nanning Zheng, and Gang Hua.
\newblock Learning disentangled classification and localization representations
  for temporal action localization.
\newblock In {\em Proceedings of the AAAI Conference on Artificial
  Intelligence}, volume~2, 2022.

\bibitem{Zhu_Wang_Tang_Liu_Zheng_Hua_2022}
Zixin Zhu, Le Wang, Wei Tang, Ziyi Liu, Nanning Zheng, and Gang Hua.
\newblock Learning disentangled classification and localization representations
  for temporal action localization.
\newblock {\em Proceedings of the AAAI Conference on Artificial Intelligence},
  36(3):3644--3652, Jun. 2022.

\bibitem{zitnik2017predicting}
Marinka Zitnik and Jure Leskovec.
\newblock Predicting multicellular function through multi-layer tissue
  networks.
\newblock {\em Bioinformatics}, 33(14):i190--i198, 2017.

\end{thebibliography}
}
\newpage

\section*{Supplementary Materials}
\begin{alphasection}
\section{Additional Details} \label{sec:introduction}

In this section, we provide some additional details of our loss functions and visualise object tracking with scene graphs for all three datasets.

\subsection{Loss Functions}
\label{sec:loss}

Our problem is therefore multi-objective, as we aim at not only recognising the label of the activity taking place but also finding its boundary (start and end time). 

Given the ground truth 
$y_s = \{ y_{s_1}, \ldots, y_{s_i}, \ldots, y_{s_N} \}$
and predicted
$\hat{y}_s = \{ \hat{y}_{s_1}, \ldots, \hat{y}_{s_i}, \ldots, \hat{y}_{s_N} \}$ activity labels,
and the ground truth
$y_{br} = \{ y_{{br}_1}, \ldots, y_{{br}_i}, \ldots, y_{{br}_N} \}$
and predicted
$\hat{y}_{br} = \{ \hat{y}_{{br}_1}, \ldots, \hat{y}_{{br}_i}, \ldots, \hat{y}_{{br}_N} \}$
temporal extent labels, our overall loss function is formulated as:
\begin{equation} 
    \mathcal{L} = \lambda \cdot \mathcal{L}_{Act} + \mathcal{L}_{Br}.
\end{equation}
The second component denotes the binary cross entropy (BCE) loss
$\mathcal{L}_{Br} = \{l_1,..., l_i,...,l_N\}^T$, where
\[
\begin{array}{l}
    l_i = - w_i \Big [ {y_{{br}_i} \cdot \log \hat{y}_{{br}_i}} + (1 - {y_{{br}_i}) \cdot \log(1 - \hat{y}_{{br}_i}}) \Big ],
\end{array}
\]
driving the recognition of the activity's temporal extent in class-agnostic manner.

$\mathcal{L}_{Act}$ denotes, instead, the \emph{BCEWithLogitsLoss} \cite{BCEWithLogitsLoss},
defined as 
$\mathcal{L}_{Act} = \{ l_{1,c}, \ldots, l_{i,c}, \ldots, l_{N,c}\}^T$, where
\begin{equation} 
\begin{aligned}
    l_{i,c} = - w_{i,c} \Big [ p_c \cdot y_{s_i,c} \cdot \log \sigma(\hat{y}_{s_i,c}) \\ + (1 - y_{s_i,c} \cdot \log (1 - \sigma(\hat{y}_{s_i,c})) \Big ].
\end{aligned}
\end{equation}
Here $i$ is the index of the sample in the batch, $c$ is the class number, $p_c$ is the weight of a positive sample for class $c$, and $\sigma$ is the sigmoid function.
This loss component is used to predict the activity labels $\{y_{s_i}\}$, and combines  classical binary cross entropy 
with a sigmoid layer\footnote{\url{https://pytorch.org/docs/stable/generated/torch.nn.BCEWithLogitsLoss.html}}. This combination is proven to be numerically stronger (as it leverages the log-sum-exp trick) than using a plain sigmoid separately followed by a BCE loss.

Finally, $\lambda$ is a weight term, which we set to the number of selected anchor proposals (128 in our case).

\begin{figure*}[h]
    \centering
    \includegraphics[width=0.75\textwidth]{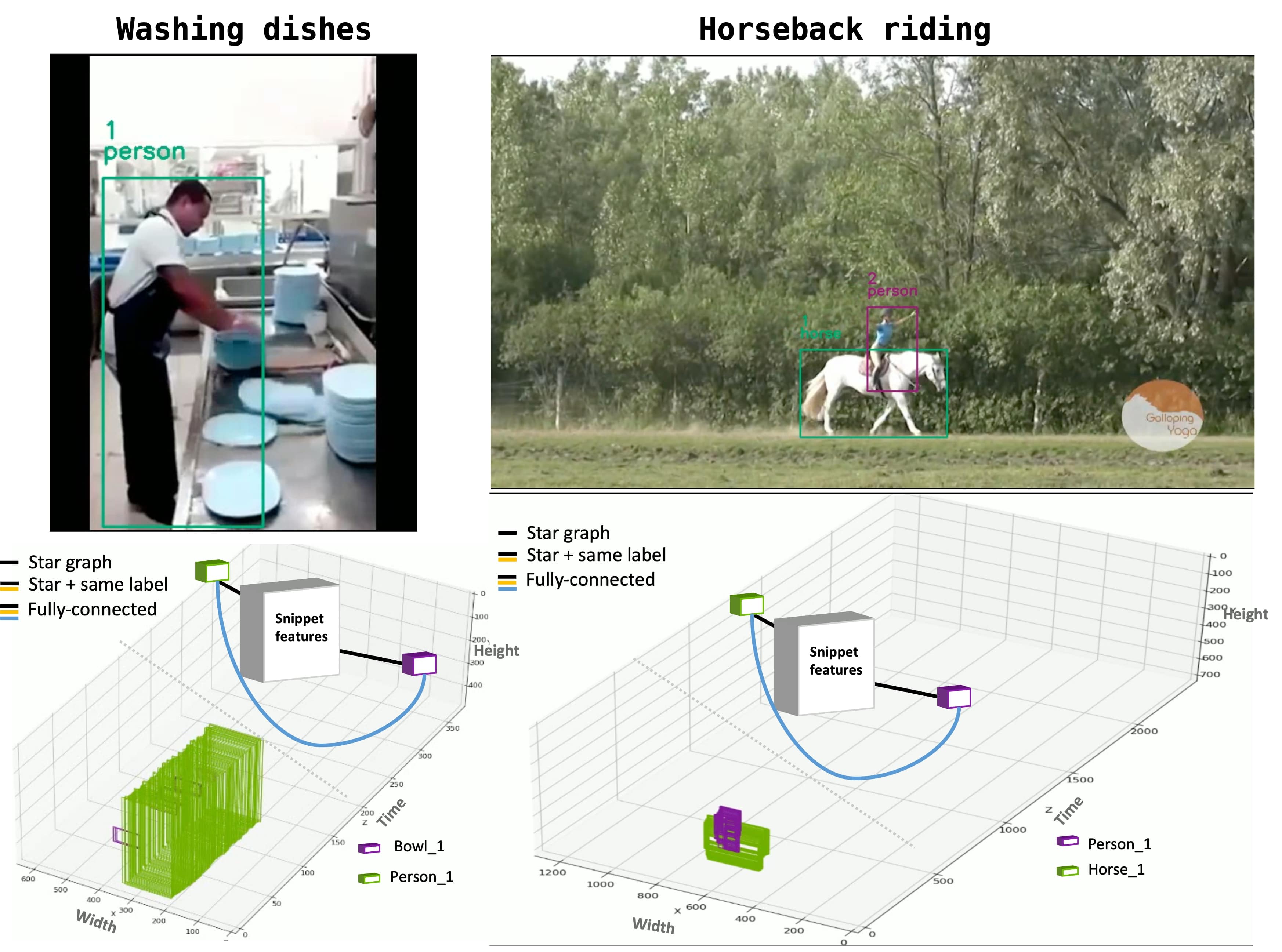}
    \caption{Visualisation of our agent detection and tracking stage using a bird’s-eye view of the spatiotemporal volume corresponding to a video segment of \emph{ActivityNet-1.3} dataset. In this figure, we present examples of activities performed by a single actor.}
    \label{fig:act_sing}
\end{figure*}

\begin{figure*}[h]
    \centering
    \includegraphics[width=0.75\textwidth]{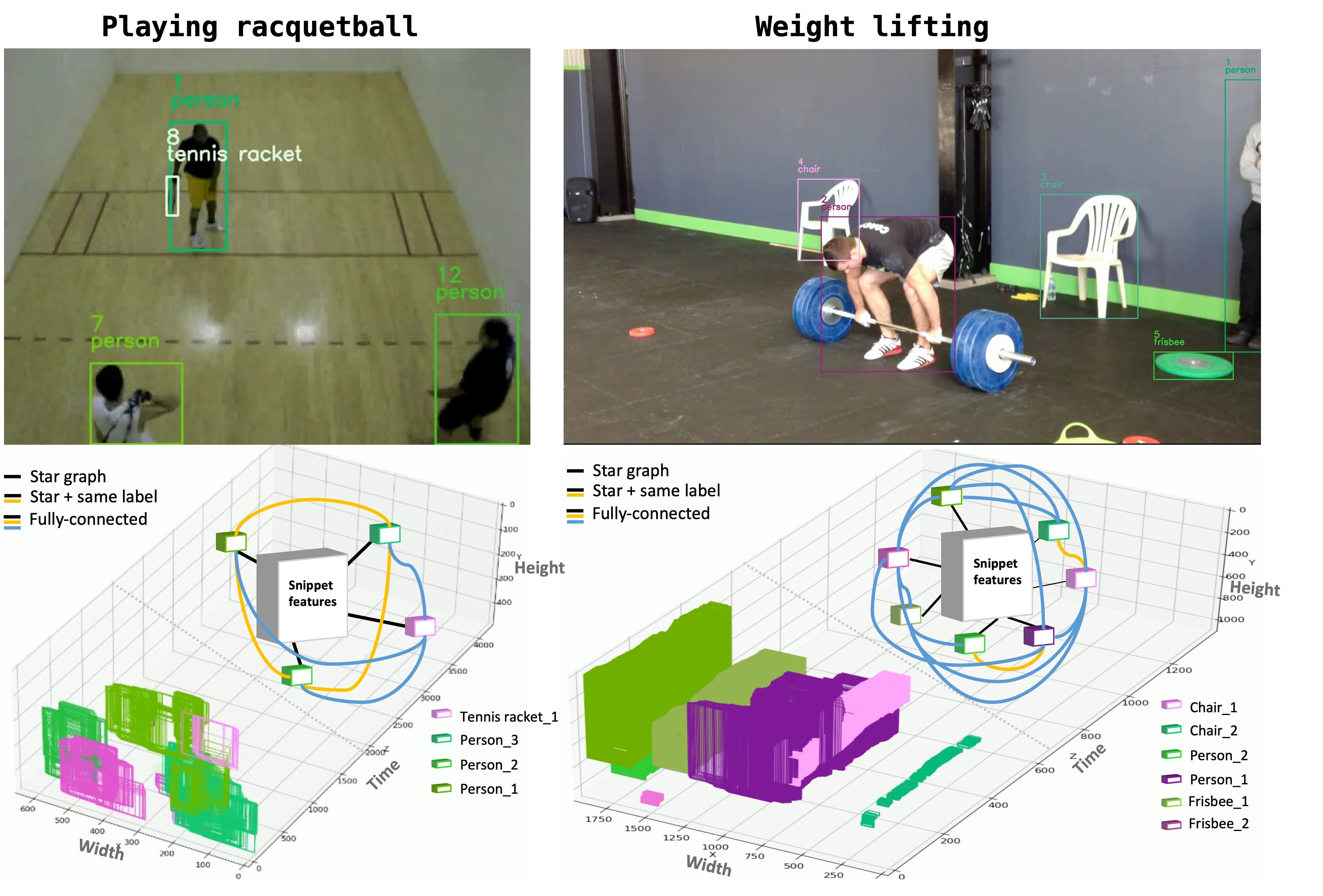}
    \caption{Visualisation of our agent detection and tracking stage using a bird’s-eye view of the spatiotemporal volume corresponding to a video segment of \emph{ActivityNet-1.3} dataset. In this figure, we show examples of activities performed by multiple actors (agent tubes).}
    \label{fig:act_multi}
\end{figure*}

\subsection{Objects Detection, Tracking, and Scene Graphs}

In addition to Fig. 2 in the paper, we also provide a pictorial illustration of agent tubes in an example video segment from all three datasets, showing both a sample frame, with overlaid the detection bounding boxes, and a bird’s-eye view of the agent tubes and a local scene graph for the snippet it belongs to.

\textbf{ActivityNet-1.3} is one of the largest temporal action localisation datasets which includes activities performed by both individual and multiple objects (agents). Therefore, we illustrate the example of both classes single agent and multi-agent activities in Fig. \ref{fig:act_sing} and \ref{fig:act_multi}, respectively.

\textbf{Thumos-14} covers different types of sports actions performed by single or mostly multiple agents. Two sample actions; baseball pitch and soccer penalty are visualised in Fig. \ref{fig:thumos}. In the figure, it can be seen that our model construct the seen graph of almost all agents despite the low-resolution videos.

\begin{figure*}[h]
    \centering
    \includegraphics[width=0.75\textwidth]{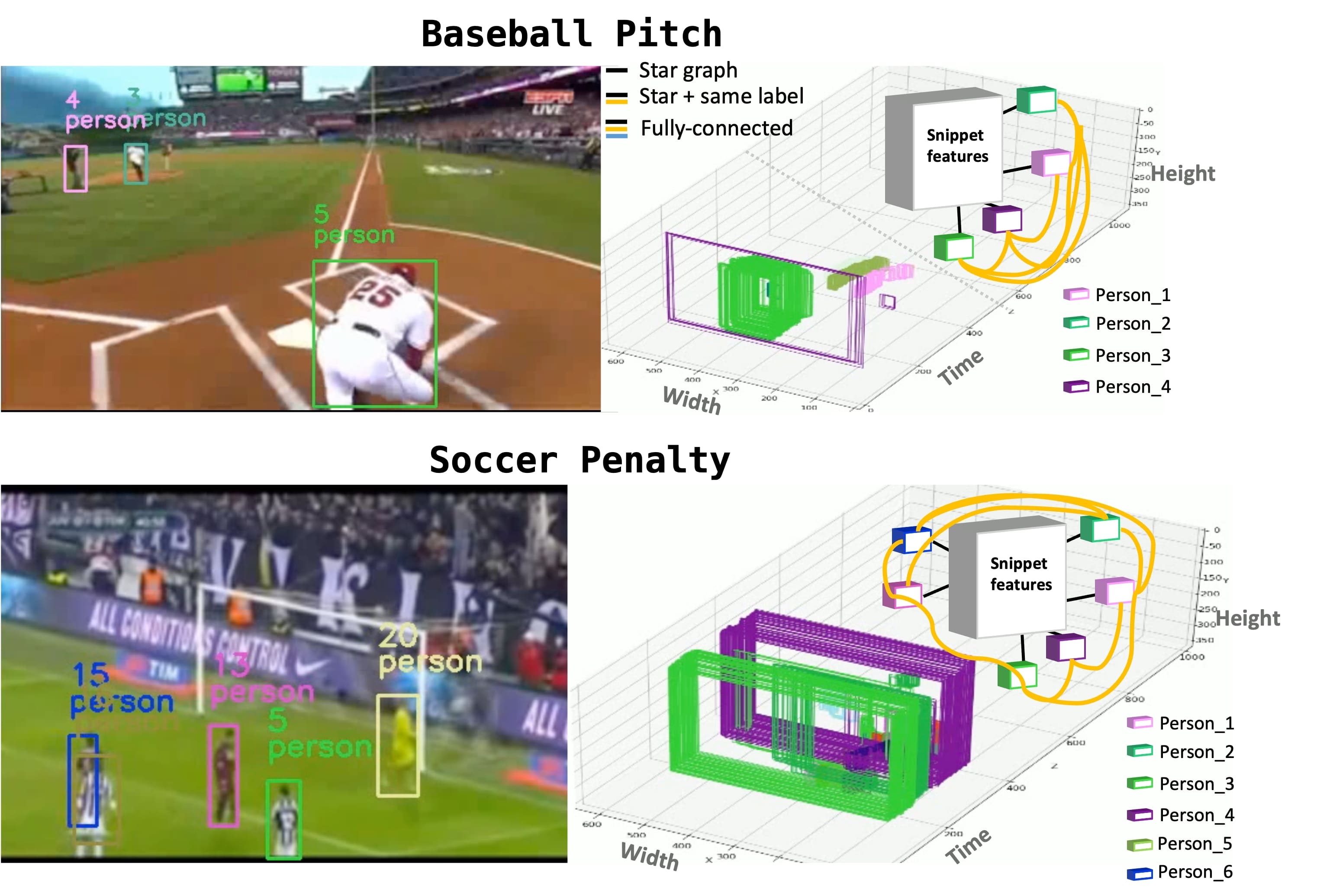}
    \caption{Visualisation of our agent detection and tracking stage using a bird’s-eye view of the spatiotemporal volume corresponding to a video segment of \emph{Thumos-14} dataset. Both of the examples show the activities performed by multiple agents (recognising sports activities).}
    \label{fig:thumos}
\end{figure*}

\textbf{ROAD}. The activities in the road dataset are mostly performed by a large number of agents, however, there are a few cases where the number of detected agents is less in number as shown in Fig. \ref{fig:road} (above). On the other hand, in Fig. \ref{fig:road} (below) we also show an example of a night video where the activity is performed by a large number of agents moving very promptly.

\begin{figure*}[h]
    \centering
    \includegraphics[width=0.75\textwidth]{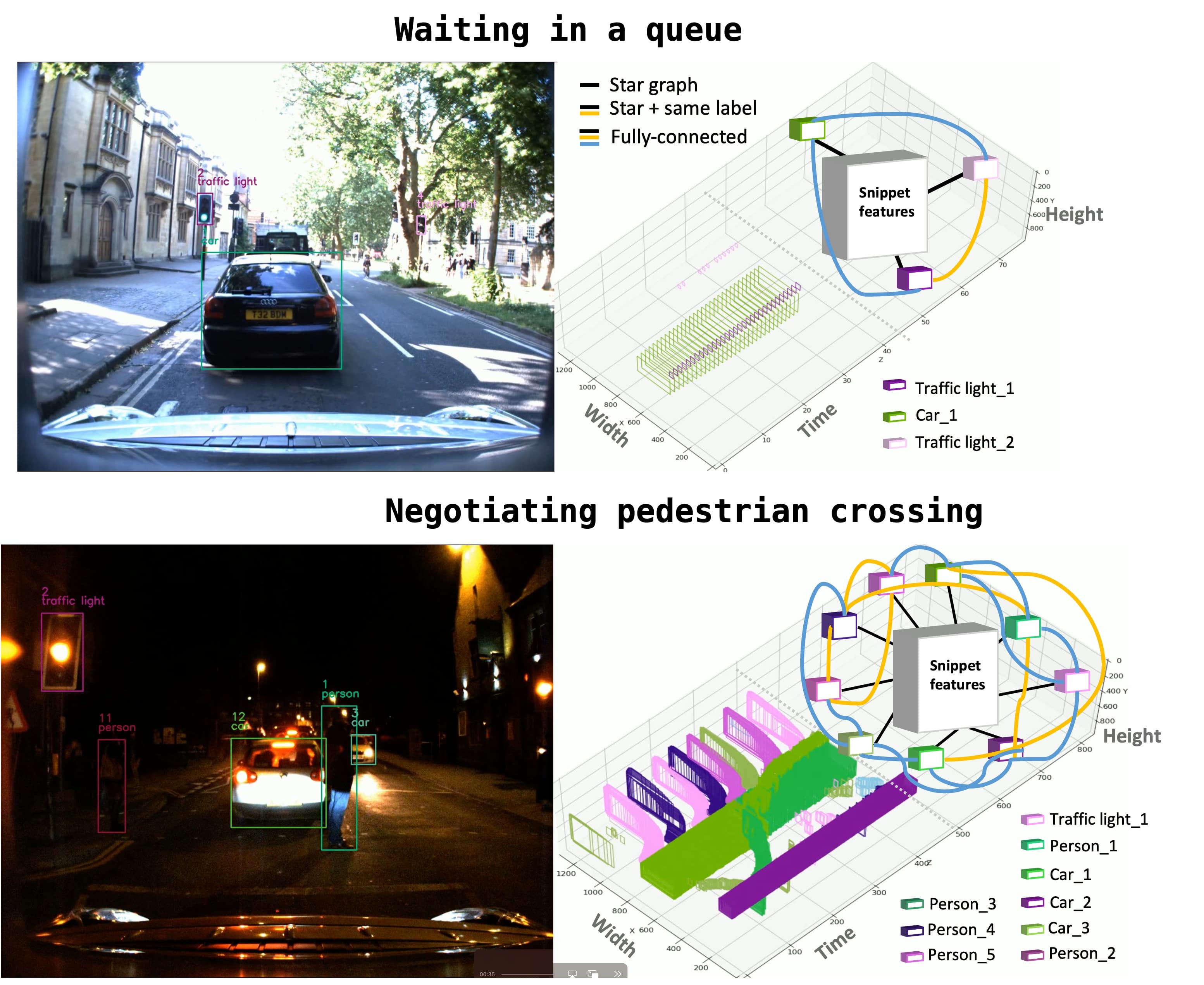}
    \caption{Visualisation of our agent detection and tracking stage
using a bird’s-eye view of the spatiotemporal volume corresponding to a video segment of the \emph{ROAD} dataset. The upper section
shows an example of activity performed by only three agents.
Below is the example of a night video where the activity is performed by multiple agents.}
    \label{fig:road}
\end{figure*}


\section{Additional Experiments}

\subsection{Qualitative Results}
We provide additional qualitative results of our method on all three datasets in Fig. \ref{fig:qaul}. The figure shows two samples per dataset, and portrays a series
of local scenes (snippets), skipping some for visualisation
purposes, with superimposed the ground truth (in green)
and the prediction of our model (in red).

\begin{figure*}[h]
    \centering
    \includegraphics[width=0.70\textwidth]{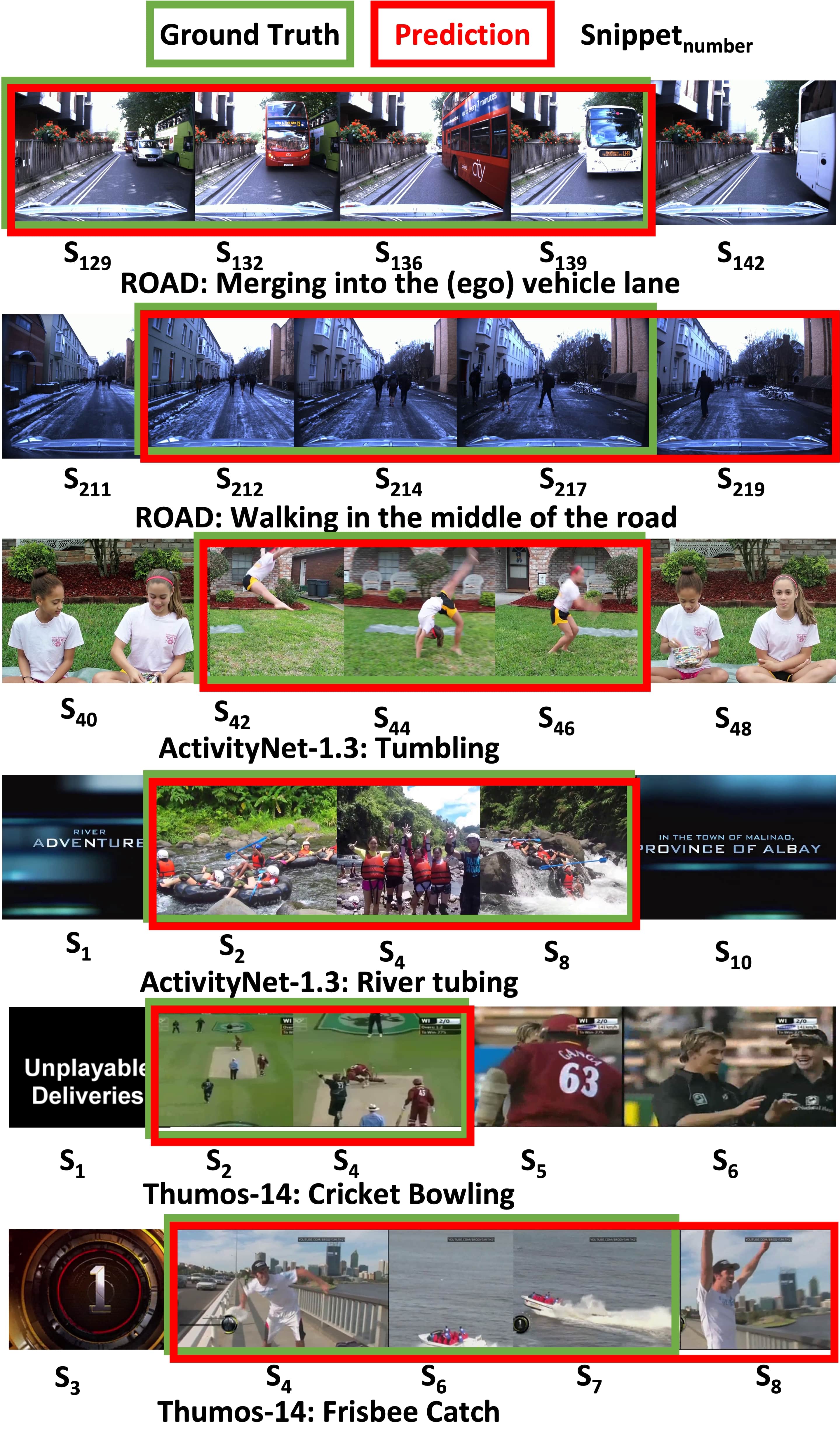}
    \caption{The qualitative results of our proposed method for all three datasets. The green rectangles covering the snippets (local scenes) are the ground truth while the yellow boxes show the prediction of our model.}
    \label{fig:qaul}
\end{figure*}
\end{alphasection}

\end{document}